\documentclass{article}

\usepackage[preprint]{neurips_2026}

\usepackage[utf8]{inputenc}
\usepackage[T1]{fontenc}
\usepackage{fix-cm}
\usepackage{hyperref}
\usepackage{url}
\usepackage{booktabs}
\usepackage{amsfonts}
\usepackage{amsmath}
\usepackage{amssymb}
\usepackage{nicefrac}
\usepackage{microtype}
\usepackage{xcolor}
\usepackage{multirow}
\usepackage{graphicx}
\usepackage{subcaption}
\usepackage{colortbl}
\usepackage{array}
\usepackage{makecell}
\usepackage{tikz}
\usepackage{pgfplots}
\pgfplotsset{compat=1.18}
\usetikzlibrary{positioning,arrows.meta,fit,backgrounds,calc}

\definecolor{aurocHigh}{HTML}{C6EFCE}    
\definecolor{aurocMid}{HTML}{FFEB9C}     
\definecolor{aurocLow}{HTML}{FFCC99}     
\definecolor{aurocRand}{HTML}{FFC7CE}    
\definecolor{aurocMiss}{HTML}{D9D9D9}    
\definecolor{artifactFlag}{HTML}{FFD700} 

\newcommand{\hi}[1]{\cellcolor{aurocHigh}{#1}}
\newcommand{\mi}[1]{\cellcolor{aurocMid}{#1}}
\newcommand{\lo}[1]{\cellcolor{aurocLow}{#1}}
\newcommand{\ra}[1]{\cellcolor{aurocRand}{#1}}
\newcommand{\ms}{\cellcolor{aurocMiss}{---}}
\newcommand{\scc}[1]{\cellcolor{artifactFlag}{#1}$^{\dagger}$}

\definecolor{driftorange}{RGB}{230,126,34}
\definecolor{driftblue}{RGB}{52,152,219}
\definecolor{driftgreen}{RGB}{39,174,96}
\definecolor{driftgray}{RGB}{236,240,241}
\definecolor{driftred}{RGB}{231,76,60}
\definecolor{driftsalmon}{RGB}{255,235,230}
\definecolor{driftmint}{RGB}{230,250,240}
\definecolor{driftyellow}{RGB}{255,245,200}
\definecolor{driftpurple}{RGB}{142,68,173}

\title{PARALLAX: Separating Genuine Progress from Benchmark Artifacts in Hallucination Detection}

\author{%
  Khizar Hussain \\
  Virginia Tech \\
  \texttt{khizar@vt.edu}
  \And
  Murat Kantarcioglu \\
  Virginia Tech \\
  \texttt{muratk@vt.edu}
}

\begin{document}

\maketitle

\begin{abstract}
Large language models (LLMs) hallucinate with confidence: their outputs can be
fluent, authoritative, and simply wrong.
In medical, legal, and scientific applications this failure causes direct harm, and
detecting it from internal model states offers a path to safer deployment.
A growing body of work reports that this problem is increasingly tractable, with
recent methods achieving high detection performance on widely used benchmarks.
We show, however, that much of this apparent progress does not survive scrutiny.
Four of the six corpora embed the ground-truth answer directly in the input prompt.
A na\"{i}ve text-similarity baseline we call \textsc{TxTemb} exploits this to
achieve near-perfect detection scores without any access to model internals.
To measure what genuine detection capability remains once these artifacts are
controlled, we conduct a large-scale evaluation spanning twenty-two detection methods,
twelve open-source models spanning six architectural families, and six corpora.
We further introduce \textbf{DRIFT}, a supervised probe over inter-layer hidden-state
transitions, as a point of comparison for live-generation detection.
Our findings suggest that the field's reported progress on hallucination detection is
substantially explained by benchmark construction artifacts in widely used corpora,
and that the majority of established baselines perform near chance under controlled
conditions; the consistent exceptions are SAPLMA and DRIFT, both supervised probes
on upper-layer hidden states.
\end{abstract}

\section{Introduction}
\label{sec:intro}

Reliable hallucination detection matters: errors in medical, legal, and scientific
applications cause direct harm~\citep{medhalsurvey2025,dahl2024legal,openai2025healthbench}, and a real-time detector requiring no extra inference
would enable safer deployment of production LLMs.
A growing body of work reports high-AUROC hallucination detectors~\citep{zhang2023siren,ji2023survey,huang2023survey}, but the field
lacks controlled cross-method comparisons and reported scores come with two serious
confounds that have not been adequately addressed.
The first is benchmark construction: popular corpora such as
HaluEval \citep{li2023halueval}, MedHallu \citep{medhallbench2024}, and
TruthfulQA \citep{lin2022truthfulqa} present the model with both a correct and a
hallucinated answer in the same prompt, so the label is correlated with surface text
differences rather than model uncertainty.
We demonstrate this with \textsc{TxTemb} (Eq.~\ref{eq:txtemb}): a TF-IDF cosine similarity score between the hallucinated and reference answer texts, requiring no access to model internals, which achieves AUROC~$= 0.98$ on HaluEval.
Any internal-state method achieving similar performance is not detecting
hallucination; it is detecting lexical phrasing patterns in the prompt.
Throughout this paper, \textsc{TxTemb} scores serve as an artifact control: a high \textsc{TxTemb} AUROC on a given corpus indicates that surface text differences, not model uncertainty, drive detection performance.
The second confound is evaluation scope: most papers evaluate one or two benchmarks
on a single model, making it impossible to distinguish a genuine mechanistic signal
from overfitting to a specific benchmark-model combination.

We address both confounds with a large-scale controlled evaluation.
Across six corpora (open-domain, medical, factual, legal, RAG, multi-domain QA),
we evaluate twenty-two detection methods on twelve instruction-tuned models spanning
six architectural families (3.8B--72B parameters).
Our six probe-based approaches include \textbf{DRIFT} (Approach E, introduced here),
self-supervised perturbation contrast (A), contrastive activation addition probe (B),
semantic entropy probe (C, adaptation of \citep{farquhar2024semantic}),
activation variance consistency (D), and answer expectation (F).
These are evaluated against sixteen established and new baselines
\citep{chen2024inside,manakul2023selfcheckgpt,su2024mind,azaria2023internal,
chuang2024dola,wang2025act,du2024haloscope,dasgupta2025hallushift,
farquhar2024semantic,fujie2024prism,niu2025hami},
with \textsc{TxTemb} artifact controls, permutation tests, and bootstrap confidence intervals applied to every method-corpus pair.
We include DRIFT among the evaluated methods; all probe hyperparameters were fixed
before the live-generation test splits (RAGTruth, HaluBench) were accessed.

Our findings raise serious concerns.
\emph{Teacher-forced} corpora embed the ground-truth answer directly in the input prompt; \emph{live-generation} corpora require the model to respond freely, with post-hoc labels.
On teacher-forced corpora, artifact-exploiting methods reach AUROC $0.85$--$1.00$ but collapse to near-chance once the artifact is removed.
On live-generation corpora, SAPLMA and DRIFT, both supervised probes, reliably exceed chance on HaluBench, reaching AUROC~$0.91$ (mean $0.879$ across twelve architectures), while every label-free method, including MIND, HaloScope, ACT, SEPs, and HalluShift, fails near chance regardless of architecture.
On RAGTruth every method falls between $0.43$ and $0.57$, supervised or otherwise.

\textbf{Our contributions:}
(1)~A large-scale controlled evaluation of 22 detection methods across 6 corpora and
12 models with \textsc{TxTemb} artifact controls, permutation tests, and bootstrap CIs.
(2)~The finding that teacher-forcing inflates reported AUROC by up to $0.43$ points,
and that SAPLMA (2023) achieves the strongest controlled HaluBench performance among
all externally published prior methods, a result obscured by artifact-contaminated
evaluations in the prior literature.
(3)~The finding that all 22 methods cluster between AUROC~$0.43$ and $0.57$ on
RAGTruth, with no method significantly exceeding chance under permutation tests,
establishing RAGTruth as a concrete unsolved benchmark for activation-based detection.
(4)~\textbf{DRIFT}, a supervised inter-layer probe within $0.004$ of SAPLMA on
HaluBench ($0.915$ vs.\ $0.911$) and statistically indistinguishable on RAGTruth
($p{=}0.82$), confirming that the positive HaluBench result is not unique to a
2023 single-layer architecture; the DRIFT-concat ablation (same layers, concatenated) further
isolates upper-layer depth selection as the primary driver of supervised probe
performance.
(5)~Code and evaluation scripts released at \url{https://anonymous.4open.science/r/llm-fuzzing-2A24}.

\begin{figure*}[!t]
\centering
\scalebox{0.82}{%
\begin{tikzpicture}[
  font=\sffamily\tiny, >=Stealth,
  box/.style={draw, rounded corners=2pt, inner sep=2pt},
  hbox/.style={box, fill=driftblue!20, draw=driftblue!60,
               minimum width=0.78cm, minimum height=0.18cm, font=\fontsize{5.5}{6.5}\selectfont\sffamily},
  phibox/.style={draw, rounded corners=2pt, fill=driftorange!12, draw=driftorange!50,
                 minimum width=2.4cm, minimum height=0.34cm, inner sep=2pt,
                 font=\fontsize{5}{6}\selectfont\sffamily, align=center},
  arr/.style={->, gray!60, thin},
  tarr/.style={->, gray!30, very thin},
]
\def\barH{4.0}

\fill[driftgray!22, draw=gray!25, rounded corners=2pt] (0.40,0.07) rectangle (1.00,\barH+0.07);
\fill[driftgray!50, draw=gray!38, rounded corners=2pt] (0.20,0.035) rectangle (0.80,\barH+0.035);
\fill[driftgray, draw=gray!55, rounded corners=2pt] (0,0) rectangle (0.6,\barH);
\foreach \y in {0.5,1.0,1.5,2.0,2.5,3.0,3.5}{
  \draw[gray!28, very thin] (0.06,\y) -- (0.54,\y);
}
\node[left=1pt, font=\fontsize{4.5}{5.5}\selectfont\sffamily, gray!50!black] at (0,0.12) {$L_0$};
\node[left=1pt, font=\fontsize{4.5}{5.5}\selectfont\sffamily, gray!50!black] at (0,\barH-0.12) {$L_n$};
\node[above, font=\sffamily\scriptsize\bfseries, gray!65!black, align=center]
  at (0.50,\barH+0.08) {LLM\\\fontsize{5}{6}\selectfont(frozen)};

\foreach \pct/\disp in {60/{0.60n}, 70/{0.70n}, 80/{0.80n}, 85/{0.85n}}{
  \pgfmathsetmacro{\yy}{\pct/100*\barH}
  \fill[driftorange!85] (-0.12,\yy-0.08) rectangle (0.72,\yy+0.08);
  \node[left=2pt, font=\fontsize{5.5}{6.5}\selectfont\sffamily\bfseries, driftorange!80!black]
    at (-0.12,\yy) {$\disp$};
}

\foreach \i/\col in {1/driftblue, 2/driftpurple, 3/driftorange,
                     4/driftblue, 5/driftpurple, 6/driftorange,
                     7/driftblue, 8/driftpurple}{
  \node[draw=\col!70!black, fill=\col!22, rounded corners=1.5pt,
        minimum width=0.22cm, minimum height=0.18cm, inner sep=0pt]
    at ({-0.42+(\i-1)*0.24}, -0.28) {};
}
\node[font=\fontsize{5}{6}\selectfont\sffamily, gray!60!black, align=center]
  at (0.50,-0.48) {prompt\,+\,response tokens};
\draw[->, gray!45, thin] (0.50,-0.16) -- (0.50,-0.08);

\node[font=\fontsize{5.5}{6.5}\selectfont\sffamily, gray!58!black, align=center]
  at (1.95, 0.44*\barH) {mean-pool\\over $T$ tokens\\$\in\mathbb{R}^d$};

\node[hbox] (h1) at (3.2, 0.60*\barH) {$h_{1}$};
\node[hbox] (h2) at (3.2, 0.70*\barH) {$h_{2}$};
\node[hbox] (h3) at (3.2, 0.80*\barH) {$h_{3}$};
\node[hbox] (h4) at (3.2, 3.60) {$h_{4}$};
\draw[arr] (0.72, 0.60*\barH) -- (h1.west);
\draw[arr] (0.72, 0.70*\barH) -- (h2.west);
\draw[arr] (0.72, 0.80*\barH) -- (h3.west);
\draw[arr] (0.72, 0.85*\barH) -- (h4.west);

\pgfmathsetmacro{\pc}{0.64*\barH}
\node[phibox] (phi34) at (6.1, {\pc+1.10})  {$\boldsymbol\phi_{34}$: $\mathbf{h}_4{-}\mathbf{h}_3,\;\cos(\cdot),\;\|\cdot\|_2\;\in\mathbb{R}^{d+2}$};
\node[font=\fontsize{9}{10}\selectfont, gray!55] (phidots) at (6.1, {\pc+0.275}) {$\vdots$};
\node[phibox] (phi12) at (6.1, {\pc-0.55}) {$\boldsymbol\phi_{12}$: $\mathbf{h}_2{-}\mathbf{h}_1,\;\cos(\cdot),\;\|\cdot\|_2\;\in\mathbb{R}^{d+2}$};

\draw[tarr] (h4.east) to[bend left=6]   (phi34.west);
\draw[tarr] (h3.east) to[out=0, in=200] (phi34.west);
\draw[tarr] (h2.east) to[out=0, in=135] (phi12.west);
\draw[tarr] (h1.east) to[out=0, in=155] (phi12.west);

\node[box, fill=gray!8, draw=gray!40, minimum width=1.6cm, minimum height=0.85cm,
      align=center, font=\fontsize{5}{6}\selectfont\sffamily]
  (zbox) at (10.2, 3.55)
  {concat $\mathbf{z}\!\in\!\mathbb{R}^{K(d+2)}$\\[1pt]
   \textcolor{driftorange!70!black}{\bfseries$49{,}164$-dim}};
\draw[tarr] (phi34.east)   to[out=0, in=175] (zbox.west);
\draw[tarr] (phidots.east) to[out=0, in=215] (zbox.west);
\draw[tarr] (phi12.east)   to[out=0, in=238] (zbox.west);

\node[box, fill=driftgreen!15, draw=driftgreen!55,
      minimum width=2.0cm, minimum height=1.2cm, align=center]
  (probe) at (10.2, 2.1)
  {{\fontsize{6.5}{7.5}\selectfont\sffamily\bfseries}LogisticProbe\\[2pt]
   \fontsize{5.5}{6.5}\selectfont L2-regularised\\
   \fontsize{5.5}{6.5}\selectfont 5-fold CV\\[1pt]
   \fontsize{5}{6}\selectfont fit on labeled pairs};
\draw[->, driftgreen!70, thick] (zbox.south) -- (probe.north);

\node[below=0.28cm of probe, align=center,
      font=\fontsize{6.5}{7.5}\selectfont\sffamily] (out)
  {\textcolor{driftred!80!black}{\bfseries$P(\text{hal.})$}\\[1pt]
   \fontsize{5}{6}\selectfont\textcolor{gray!60!black}{$w_E=$ contrastive dir.}};
\draw[->, driftgreen!70, thick] (probe.south) -- (out.north);

\node[draw=gray!35, dashed, rounded corners=2pt, fill=gray!5, anchor=west,
      text width=5.5cm, inner sep=3pt,
      font=\fontsize{5.5}{6.5}\selectfont\sffamily, align=left]
  at (1.5, 0.55)
  {\textbf{SAPLMA:} single layer, last token $\to\mathbb{R}^d\to$ probe.\quad
   \textbf{HalluShift:} scalar $1{-}\cos(\mathbf{h}_a,\mathbf{h}_b)$.\quad
   \textbf{DRIFT-concat:} 3-layer concat, no $\boldsymbol\phi$ diffs.};
\end{tikzpicture}%
}
\vspace{-6pt}
\caption{{\footnotesize\textbf{DRIFT architecture.}
Hidden states are tapped at four upper-layer positions (60--85\% depth), mean-pooled over $T$ tokens,
and differenced across all $C(4,2){=}6$ pairs to form $\boldsymbol\phi_{ab}\in\mathbb{R}^{d+2}$,
then concatenated into $\mathbf{z}\in\mathbb{R}^{49{,}164}$.
An L2-regularised logistic probe maps $\mathbf{z}$ to $P(\text{hallucination})$; its weight vector
$w_E$ doubles as a hallucination contrastive direction. Dashed box: baseline comparisons.}}
\label{fig:drift-architecture}
\end{figure*}
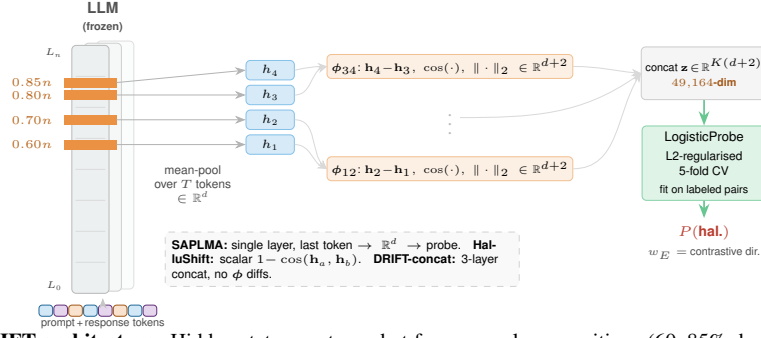

\section{Related Work}
\label{sec:related}
\vspace{-5pt}

\noindent\textbf{Internal-state hallucination probes.}
SAPLMA~\citep{azaria2023internal} trains a logistic probe on the last-layer hidden state at the final token.
INSIDE~\citep{chen2024inside} uses eigenvalues of the hidden-state covariance matrix; MIND~\citep{su2024mind} applies an unsupervised variance detector at runtime; HaloScope~\citep{du2024haloscope} scores reconstruction error from a truncated SVD; HaMI~\citep{niu2025hami} uses multiple-instance learning over stochastic samples; PRISM~\citep{fujie2024prism} unifies internal-state signals via cross-attention; HalluShift~\citep{dasgupta2025hallushift} computes a scalar inter-layer cosine distance; and ACT-ViT~\citep{barshalom2025actvit} applies activation tensor decomposition via vision transformer encoders.
DRIFT is an alternative in this family that extends the probing approach to inter-layer directions, building on evidence that models encode incorrectness before it appears in output~\citep{orgad2025llmsknow,ferrando2025entity,quevedo2024tokenlevel,hallucinationlayers2025,liang2025neuralprobe} and on probing methodology foundations~\citep{belinkov2022probing,tenney2019bert}.

\noindent\textbf{Benchmark artifacts and output-space methods.}
HaluEval~\citep{li2023halueval} concatenates correct and hallucinated answers in the same prompt, enabling lexical shortcuts unrelated to model uncertainty; we quantify this via the \textsc{TxTemb} gap criterion. HADES~\citep{liu2022hades} provides a reference-free token-level alternative, though surface artifacts remain pervasive across popular corpora.
Analogous annotation challenges arise in faithfulness evaluation~\citep{maynez2020faithfulness,laban2022summac,slobodkin2023summac,min2023factscore} and knowledge-limit studies~\citep{petroni2019language,mallen2022entity} further motivate live-generation corpora.
SelfCheckGPT~\citep{manakul2023selfcheckgpt} measures NLI inconsistency across $N{=}10$ samples; P(True)~\citep{kadavath2022know} prompts the model to self-assess; DoLa~\citep{chuang2024dola} contrasts early/late-layer logits; and semantic entropy~\citep{farquhar2024semantic,kuhn2023semantic} clusters responses by entailment.
All require extra inference passes and, as we show, yield near-chance AUROC on RAGTruth.

\noindent\textbf{Activation steering.}
CAA~\citep{panickssery2023steering}, RepE~\citep{zou2023representation}, ITI~\citep{li2023iti}, ACT~\citep{wang2025act}, FASB~\citep{cheng2025fasb}, CCS~\citep{burns2023discovering}, and ActAdd~\citep{turner2023activation} build and inject contrastive steering vectors.
DRIFT's probe weight vector doubles as a hallucination contrastive direction, offering a unified detect-then-steer path without a separate steering-vector computation.
\citet{balepur2025singledirection} show no single direction generalises across prompts, motivating multi-pair designs; dynamic and real-time variants~\citep{yin2025dynamic,obeso2025realtime,repengsurvey2025} operationalise these at deployment.

\noindent\textbf{Mechanistic interpretability, cascades, and domain evaluation.}
Residual stream analysis~\citep{elhage2021mathematical,meng2022rome} and hallucination studies in RAG and reasoning~\citep{sun2024redeep,dassen2026factum,yang2025reasoning} provide mechanistic grounding for internal-state methods; hallucinated tokens propagate autoregressively~\citep{zhang2024snowball,dziri2023faith,lee2022factuality}.
Domain evaluation in clinical and high-stakes settings~\citep{asgari2025creola,openai2025healthbench,wang2025cares,perez2022red,ness2024medfuzz} extends the landscape.
\vspace{-10pt}
\section{Experimental Setup}
\label{sec:setup}
\vspace{-6pt}
\noindent\textbf{Models.}
We evaluate 12 instruction-tuned~\citep{ouyang2022rlhf} models spanning six families (3.8B--72B parameters):
Llama (3.1-8B, 3.3-70B)~\citep{touvron2023llama2,dubey2024llama3}, Qwen (3-8B, 3-32B, 2.5-72B), DeepSeek-R1 distillations
(7B, 70B; reasoning models with chain-of-thought~\citep{wei2022chain}), Mistral (Mixtral-8x7B MoE, Mistral-24B), Gemma (3-4b, 3-12b),
and Phi-4-mini.
Llama-3.3-70B is the primary model (all 22 methods); the remaining 11 run DRIFT, SAPLMA,
and SelfCheckGPT-NLI on TruthfulQA and HaluBench. Full model identifiers appear in Appendix~\ref{app:models}.

\noindent\textbf{Train/test split.}
All corpora are split 80/20 stratified by label (test set strictly held out); cross-validation methods further split the training set into 5 folds.

\noindent\textbf{Compute.}
Full evaluation takes 4--6~h per model for 8B and 18--24~h for 70B (Appendix~\ref{app:reproducibility}).
INSIDE is the slowest baseline ($\approx$6~h per corpus on 70B).

\noindent\textbf{Computational cost per method.}
All probe-based methods share one hidden-state extraction pass per prompt ($15.6$\,s
loading, $9.6$\,ms/prompt amortised).
At test time: DRIFT-logp adds $2.1$\,ms, SAPLMA $111$\,ms, DRIFT $1.2$\,s.
SelfCheckGPT-NLI requires ten additional GPU inference passes per prompt,
placing it in a wholly different cost class.
Full per-method benchmarks are given in Appendix~\ref{app:compute} (Table~\ref{tab:compute}) and the Pareto frontier is presented in Appendix~\ref{app:pareto}.

\subsection{Benchmark Suite}
\label{sec:corpora}
\vspace{-5pt}

We evaluate on six publicly available corpora spanning domain, annotation style,
and generation format (Table~\ref{tab:corpora}).

\begin{table}[t]
\centering
\vspace{-8pt}
\caption{\scriptsize Benchmark suite. \textbf{TF} = teacher-forced (answer text in prompt);
\textbf{LG} = live-generated (model produces the response freely).}
\label{tab:corpora}
\resizebox{\linewidth}{!}{%
\begin{tabular}{llrllc}
\toprule
\textbf{Corpus} & \textbf{Domain} & \textbf{N} & \textbf{Balance} &
\textbf{Label source} & \textbf{Format} \\
\midrule
HaluEval~\citep{li2023halueval}      & Open-domain QA  & 500   & 250/250 & ChatGPT-generated pairs~\citep{achiam2023gpt4} & TF \\
MedHallu~\citep{medhallbench2024}    & Medical QA      & 500   & 250/250 & Reference pairs           & TF \\
TruthfulQA~\citep{lin2022truthfulqa} & Factual QA      & 1,634 & 817/817 & Reference pairs           & TF \\
Legal~\citep{dahl2024legal}          & Legal case QA   & 499   & 159/340 & Expert correctness score  & TF \\
RAGTruth~\citep{niu2024ragtruth}     & RAG (news/QA)   & 500   & 153/347 & Expert annotation         & LG \\
HaluBench~\citep{ravi2024lynx}       & Multi-domain QA & 500   & 250/250 & PatronusAI PASS/FAIL      & LG \\
\bottomrule
\end{tabular}}
\end{table}

\begin{figure}[t]
\centering
\scalebox{0.82}{%
\begin{tikzpicture}[font=\sffamily\tiny, >=Stealth]
\node[draw=driftred!40, rounded corners=3pt, fill=driftsalmon,
      minimum width=5.6cm, minimum height=2.8cm] (LP) at (0,0) {};
\node[anchor=north, font=\sffamily\scriptsize\bfseries, text=driftred!75!black]
  at ([yshift=-0.16cm]LP.north) {Teacher-Forced};
\node[anchor=north, font=\fontsize{5.5}{6.5}\selectfont\sffamily, gray!65!black]
  at ([yshift=-0.40cm]LP.north) {HaluEval $\cdot$ MedHallu $\cdot$ Legal $\cdot$ TruthfulQA};
\node[draw=gray!40, rounded corners=1pt, fill=white,
      font=\ttfamily\fontsize{5.5}{6.5}\selectfont, inner sep=2pt, align=left, anchor=north,
      text width=5.0cm]
  (lbox) at ([yshift=-0.82cm]LP.north)
  {prompt + ``[Correct: X]\ [Hallucination: Y]''\\$\rightarrow$ label embedded in text};
\node[anchor=north west, font=\fontsize{5.5}{6.5}\selectfont\sffamily, text width=5.2cm]
  at ([yshift=-0.08cm, xshift=0.12cm]lbox.south west)
  {\textbf{TxTemb AUROC:} 0.69--0.98\quad\textbf{R\textsuperscript{2}:} 0.81};
\node[draw=none, rounded corners=2pt, fill=driftred,
      font=\fontsize{5.5}{6.5}\selectfont\sffamily\bfseries, text=white, inner sep=2pt,
      anchor=south] at ([yshift=0.12cm]LP.south)
  {$\triangle$\ Artifact: answer text in prompt};
\node[draw=driftgreen!40, rounded corners=3pt, fill=driftmint,
      minimum width=5.6cm, minimum height=2.8cm] (RP) at (6.2,0) {};
\node[anchor=north, font=\sffamily\scriptsize\bfseries, text=driftgreen!55!black]
  at ([yshift=-0.16cm]RP.north) {Live-Generation};
\node[anchor=north, font=\fontsize{5.5}{6.5}\selectfont\sffamily, gray!65!black]
  at ([yshift=-0.40cm]RP.north) {RAGTruth $\cdot$ HaluBench};
\node[draw=gray!40, rounded corners=1pt, fill=white,
      font=\ttfamily\fontsize{5.5}{6.5}\selectfont, inner sep=2pt, align=left, anchor=north,
      text width=5.0cm]
  (rbox) at ([yshift=-0.82cm]RP.north)
  {prompt $\to$ model generates freely\\$\to$ post-hoc expert label};
\node[anchor=north west, font=\fontsize{5.5}{6.5}\selectfont\sffamily, text width=5.2cm]
  at ([yshift=-0.08cm, xshift=0.12cm]rbox.south west)
  {\textbf{TxTemb AUROC:} $\approx$0.50\quad\textbf{R\textsuperscript{2}:} 0.08};
\node[draw=none, rounded corners=2pt, fill=driftgreen!65!black,
      font=\fontsize{5.5}{6.5}\selectfont\sffamily\bfseries, text=white, inner sep=2pt,
      anchor=south] at ([yshift=0.12cm]RP.south)
  {\checkmark\ Real signal: model uncertainty};
\end{tikzpicture}%
}

\caption{{\scriptsize\textbf{Two evaluation regimes.}
\emph{Left:} Teacher-forced benchmarks embed the answer in the prompt;
\textsc{TxTemb} achieves AUROC 0.69--0.98 and explains 81\% of variance
in method AUROC ($R^2{=}0.81$), a formatting artifact.
\emph{Right:} On live-generation benchmarks \textsc{TxTemb} is at chance
(${\approx}0.50$, $R^2{=}0.08$), isolating genuine internal-state signal.
}}
\label{fig:two-regime}
\end{figure}
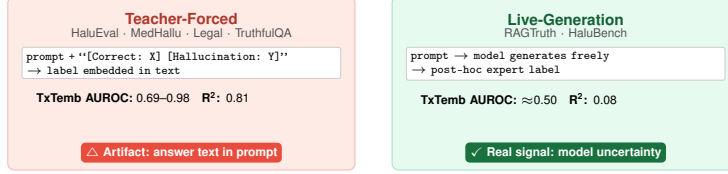

Four corpora use teacher-forced (TF) construction, where both a correct and a
hallucinated response appear in the prompt; the remaining two (RAGTruth, HaluBench)
are live-generated (LG), with the model responding freely.
The TF artifact is strongest on HaluEval (\textsc{TxTemb}~$=0.98$), moderate on MedHallu
(\textsc{TxTemb}~$=0.71$) and Legal (\textsc{TxTemb}~$=0.89$; despite nuanced expert correctness judgements,
surface-text cues remain highly predictive),
and weakest on TruthfulQA (\textsc{TxTemb}~$=0.69$, which still requires factual knowledge
to distinguish correct from incorrect answers).
Figure~\ref{fig:two-regime} illustrates the two regimes.

RAGTruth \citep{niu2024ragtruth} is our primary live-generation benchmark:
responses are generated by a RAG pipeline~\citep{lewis2020rag} under normal inference conditions, where faithfulness to source passages is essential~\citep{shi2023trusting}, and
annotated post-hoc by human experts; no answer text is embedded in the prompt.
HaluBench \citep{ravi2024lynx} uses PatronusAI's automated LLM-as-judge evaluator~\citep{zheng2023judging} on freely generated multi-domain QA;
its absence of answer text in the prompt gives it the similar live-generation character.

\noindent\textbf{Artifact control: text-embed baseline.}\label{sec:artifact_ctrl}
On teacher-forced corpora, hallucinated and correct answers differ in surface text.
We implement \textsc{TxTemb}: a purely text-based detection signal defined as
\begin{equation}
  s_{\textsc{TxTemb}}(r) = \cos\!\bigl(\mathrm{tfidf}(r_{\mathrm{hal}}),\,\mathrm{tfidf}(r_{\mathrm{ref}})\bigr),
  \label{eq:txtemb}
\end{equation}
where $r_{\mathrm{hal}}$ and $r_{\mathrm{ref}}$ are the hallucinated and reference answer texts.
The key insight is that \textsc{TxTemb} accesses only the text of the responses, with no knowledge of the model's internal states, probability distributions, or reasoning process.
If an internal-state method achieves AUROC comparable to \textsc{TxTemb} on the same corpus, its detection signal is explained by the same surface-text differences that a lexical baseline can already exploit.
In that case, the method is not detecting genuine model uncertainty or hallucination-prone internal states; it is, at best, re-encoding information already present in the prompt text through hidden representations.
Any performance claim for such a method on that corpus is therefore attributable to the benchmark construction artifact rather than to a meaningful internal-state signal.

A cell is flagged ($\dagger$) if the method's AUROC is within $0.05$ of \textsc{TxTemb} AUROC on the same corpus.
The $0.05$ threshold is chosen to exceed typical bootstrap confidence interval half-widths observed in our experiments (mean CI half-width: $0.024$), ensuring that a flagged method's performance advantage over the surface baseline is not statistically distinguishable from zero.
This is a deliberately conservative criterion: a method that outperforms \textsc{TxTemb} by only $0.05$ AUROC points may still be exploiting the same artifact through its hidden representations, and we do not wish to overstate the absence of artifact contamination.

\noindent\textbf{Evaluation metrics and verification.}
We use Area Under the ROC Curve (AUROC)~\citep{fawcett2006roc} as our \textit{primary metric} because it is threshold-independent, invariant to class imbalance, and measures the probability that a randomly chosen positive instance is ranked above a randomly chosen negative one, making it appropriate for comparing detectors across the six corpora in our benchmark, which vary substantially in label balance (Table~\ref{tab:corpora}).

For all cells with AUROC~$> 0.85$ we additionally compute:
(i)~bootstrap 95\% CI ($n=1000$ resamples);
(ii)~permutation-test null mean ($n=30$ label shuffles);
(iii)~\textsc{TxTemb} gap.
A cell is declared \emph{Validated} if the CI lower bound exceeds $0.80$
\emph{and} the \textsc{TxTemb} gap exceeds $0.05$.

\section{Methods}
\label{sec:methods}
\vspace{-5pt}
Our evaluation covers six detection approaches of our own design alongside fourteen external baselines spanning label-free, unsupervised, decoding-time, and supervised-probe paradigms.
\textbf{DRIFT} (Approach~E) is our primary design contribution and the strongest performer.
Approaches A and E are self-supervised (no labels), C and D require $N{=}10$ stochastic samples per prompt, and B and F require teacher-forced format.
All six share a common feature backbone: hidden states from a single forward pass pooled per-layer across four upper-layer checkpoints (Section~\ref{sec:infrastructure}), enabling any subset to be evaluated on the same stored activations.

\noindent\textbf{Compute and quantisation.}\label{sec:infrastructure}
All experiments run on a server with $8\times$ NVIDIA L40S GPUs (46\,GB each).
Hidden-state extraction runs on a single GPU in 4-bit NF4 quantisation (bitsandbytes double-quantisation) to fit 70B models in memory; batch generation uses 4-GPU tensor parallelism (\texttt{tp=4}).
One model--corpus pair (Qwen3-32B $\times$ TruthfulQA) required re-extraction with 8-bit quantisation after NF4 produced NaN activations (KV-cache overflow at layers 38, 44, 51, 54); the resolved result (DRIFT AUROC $= 0.671$) is included in Table~\ref{tab:multimodel}.

\noindent\textbf{Layer selection.}
We extract hidden states at fractional depths $\{0.60,\,0.70,\,0.80,\,0.85\} \times L$ (Llama-3.3-70B: layers 48, 56, 64, 68), confirmed by Appendix~\ref{app:layer_ablation}.
Truth-relevant directions reside in the upper network third~\citep{marks2023geometry} (AUROC 0.741 at 20\% vs.\ 0.898 at 85\%), final layers are dominated by next-token prediction, and four taps outperform any single tap by $+0.017$ AUROC.
These fractions were fixed before seeing any results and applied uniformly across all twelve architectures.

\noindent\textbf{Hidden state extraction.}
A forward pass over $[\text{prompt} + \text{response}]$ yields hidden states at response
token positions. Let $T$ denote the number of response tokens and $\mathbf{h}^{(\ell)}_t \in \mathbb{R}^d$
the hidden state at layer $\ell$ and position $t$. The per-layer representation is mean-pooled:
\vspace{-4pt}
\begin{equation}
  \mathbf{h}^{(\ell)} = \frac{1}{T}\sum_{t=1}^{T} \mathbf{h}^{(\ell)}_t \;\in\; \mathbb{R}^{d}.
  \label{eq:pool}
\end{equation}
\vspace{-6pt}

\noindent The four layer vectors are concatenated to form $\mathbf{h} \in \mathbb{R}^{4d}$.
The full prompt is included so activations are contextualised on the question.

\noindent\textbf{Probe architectures.}
Two probe types are used: a \emph{LogisticProbe} (L2-regularised logistic regression, 5-fold CV over $C\!\in\!\{0.001,0.01,0.1,1\}$) and an \emph{MLPProbe} (two-layer feedforward, $\min(256,d/4)$ hidden units, ReLU, BCE loss). The LogisticProbe weight vector simultaneously serves as the decision boundary and the hallucination contrastive direction~\citep{li2023iti}.

\noindent\textbf{Approach~A: Perturbation~Contrast (Self-Supervised).}
Four strategies generate a corrupted response $r'$ from prompt $p$ and response $r$
(entity swap, numerical corruption, negation flip, boundary violation).
An MLPProbe is trained on $\Delta\mathbf{h} = \mathbf{h}(p,r) - \mathbf{h}(p,r')$.
No hallucination labels are required; the perturbation provides supervision.

\noindent\textbf{Approach~B: Contrastive~Activation Addition.}
For teacher-forced data with paired correct/hallucinated responses, we compute
$\mathbf{v}_\text{CAA} = \frac{1}{N}\sum_i \mathbf{h}(p_i, r^+_i) - \mathbf{h}(p_i, r^-_i)$
and score $s_B = \cos(\mathbf{h}(p,r),\mathbf{v}_\text{CAA})$.
$^\dagger$~AUROC~$=1.0$ on MedHallu is an artifact of the teacher-forced construction.
Approach~B requires paired responses for the same prompt (teacher-forced only). In MedHallu, GPT-4 generated pairs with the reference answer visible in the prompt, so $\mathbf{v}_\text{CAA}$ encodes surface-text agreement rather than a genuine hallucination signal, and the method does not apply to live-generation corpora.

\noindent\textbf{Approach~C: Semantic~Entropy Probe.}
Semantic entropy $H=-\sum_c p_c\log p_c$ is computed from $N{=}10$ stochastic samples
\citep{farquhar2024semantic}, then an MLPProbe is trained on $\mathbf{h}$ to predict $H$.
This amortises entropy computation to a single inference pass.

\noindent\textbf{Approach~D: Activation~Variance Consistency.}
Element-wise variance $\mathbf{v}=\operatorname{Var}(\mathbf{h}_1,\ldots,\mathbf{h}_N)$
over $N{=}10$ stochastic completions is fed to a LogisticProbe, capturing which
activation dimensions are uncertain across samples.

\noindent\textbf{Approach~E: DRIFT.}
Let $\mathcal{L} = \{l_1,l_2,l_3,l_4\}$ denote the four selected layers and
$\binom{\mathcal{L}}{2}$ the set of $K{=}6$ unordered layer pairs.
For each pair $(l_a, l_b)$ with $l_a < l_b$, the per-pair feature block is:
\begin{equation}
  \boldsymbol{\phi}_{ab} = \Bigl[\,
    \mathbf{h}^{(l_b)} - \mathbf{h}^{(l_a)},\;
    \cos\!\bigl(\mathbf{h}^{(l_a)},\mathbf{h}^{(l_b)}\bigr),\;
    \bigl\|\mathbf{h}^{(l_b)} - \mathbf{h}^{(l_a)}\bigr\|_2
  \,\Bigr] \;\in\; \mathbb{R}^{d+2},
  \label{eq:phi}
\end{equation}
and the full DRIFT feature vector is the concatenation over all pairs:
\begin{equation}
  \mathbf{z} = \Bigl[\boldsymbol{\phi}_{ab}\Bigr]_{(l_a,\,l_b)\,\in\,\binom{\mathcal{L}}{2}}
  \;\in\; \mathbb{R}^{K(d+2)}.
  \label{eq:drift_feat}
\end{equation}
For Llama-3.3-70B ($d{=}8192$, $K{=}6$) this gives $6\times8194=49{,}164$ dimensions.
A StandardScaler + LogisticProbe is fitted on $\mathbf{z}$.
This extends HalluShift's scalar cosine distance to a directional, multi-layer
probed representation: the probe learns \emph{which inter-layer directions}
are diagnostic, not just that some change occurred.
See Figure~\ref{fig:drift-architecture} for an overview.
\textbf{DRIFT-concat} is a controlled ablation of Approach~E:
it uses the same three tap layers ($l \in \{0.60n,\,0.70n,\,0.80n\}$) but mean-pools and concatenates
the hidden states directly without inter-layer difference features,
isolating the contribution of the inter-layer differencing.

\noindent\textbf{Approach~F: Answer~Expectation.}
A LogisticProbe is trained on $[\mathbf{h}_\text{after}{-}\mathbf{h}_\text{before},\;
\cos(\mathbf{h}_\text{before},\mathbf{h}_\text{after}),\;
\|\mathbf{h}_\text{before}{-}\mathbf{h}_\text{after}\|]$,
where states are extracted just before and after the answer token.
$^\dagger$~Teacher-forced artifact: $\mathbf{h}_\text{before}$ encodes the answer text.
\textbf{DRIFT-logp} trains a logistic classifier on
token log-probability time series statistics (mean, min, variance, slope, entropy)
over the response; it requires no hidden-state access and adds only 2\,ms latency.

\noindent\textbf{Baselines.}\label{sec:baselines}
We evaluate fourteen external baselines spanning four paradigms.
\textbf{Label-free, multi-sample:} \textbf{INSIDE} \citep{chen2024inside} scores the mean log-eigenvalue of the hidden-state covariance across $N$ response samples;
\textbf{SelfCheckGPT-NLI} \citep{manakul2023selfcheckgpt} measures NLI contradiction across $N{=}10$ samples (official code);
\textbf{MIND} \citep{su2024mind} trains an MLP using entropy-derived pseudo-labels (reimplemented with entropy thresholding).
\textbf{Unsupervised representations:} \textbf{HaloScope} \citep{du2024haloscope} scores reconstruction error from a truncated SVD of training activations;
\textbf{HalluShift} \citep{dasgupta2025hallushift} uses the scalar inter-layer cosine distance $1{-}\cos(\mathbf{h}^{(0.6L)}, \mathbf{h}^{(0.85L)})$ (Approach~E is the learnable extension);
\textbf{DoLa} \citep{chuang2024dola} contrasts early/late-layer logits at decoding time.
\textbf{Decoding-time signals:} \textbf{ACT} \citep{wang2025act} fits logistic regression on concatenated hidden states;
\textbf{SEPs} \citep{farquhar2024semantic} predicts semantic entropy from hidden states;
\textbf{IRIS}$^\star$ uses the model's self-verification probability $P(\text{``No''}) {-} P(\text{``Yes''})$ \citep{kadavath2022know};
\textbf{log-prob} (mean response token log-probability) and \textbf{P(True)}~\citep{kadavath2022know} (self-assessment prompt) require no hidden states.
\textbf{Supervised probes:} \textbf{SAPLMA} \citep{azaria2023internal} uses single-layer last-token logistic regression with cross-validated layer selection;
\textbf{PRISM} \citep{fujie2024prism} uses a truth-prompting forward pass to extract a classification embedding;
\textbf{HaMI} \citep{niu2025hami} applies multiple-instance learning over stochastic response samples.

\section{Results}
\label{sec:results}
\vspace{-6pt}
\subsection{Main Results: Llama-3.3-70B}
\label{sec:main_results}
\vspace{-5pt}
We report results on the primary evaluation model, Llama-3.3-70B, before
presenting the cross-model sweep and artifact analysis.
Two patterns dominate the table: on four of six corpora, the reported numbers
look strong ($> 0.85$), but the artifact controls reveal that most of this
performance is not mechanistically meaningful.
On the two corpora where teacher-forcing contamination is minimal (TruthfulQA and
the verified HaluBench cells), DRIFT and SAPLMA consistently lead the field,
with SAPLMA narrowly ahead on TruthfulQA ($0.75$ vs.\ $0.73$) and both methods tied on HaluBench ($0.91$).

Table~\ref{tab:main_results} reports all 22 methods; verification details (bootstrap CIs, permutation tests) appear in Appendix~\ref{app:verification}.
Cells are colour-coded: \colorbox{aurocHigh}{green} ($\geq 0.85$),
\colorbox{aurocMid}{yellow} ($0.70$--$0.84$),
\colorbox{aurocLow}{orange} ($0.55$--$0.69$),
\colorbox{aurocRand}{red} ($< 0.55$);
\colorbox{artifactFlag}{gold}$^\dagger$ = teacher-forcing artifact detected.
Bold marks the best artifact-free result per column.

\begin{table}[t]
\centering
\vspace{-8pt}
\caption{AUROC on Llama-3.3-70B for all 22 methods, grouped by detection paradigm.
$^\dagger$~artifact detected: \textsc{TxTemb} gap $< 0.05$.
$^\ddagger$~AVC inverted on TruthfulQA (confident myths have low variance; flipped AUROC\,$=0.73$).
\textbf{Bold} = best artifact-free result per corpus.
Output/logit methods require additional LLM inference at test time.
``---'' = not applicable (B, F require teacher-forced format).
$^\star$~our implementation of concept from \citet{kadavath2022know}; no public code available.}
\label{tab:main_results}
\setlength{\tabcolsep}{2.5pt}
\scriptsize
\begin{tabular}{lcccccc}
\toprule
\textbf{Method} &
\textbf{HaluEval} & \textbf{MedHallu} & \textbf{TruthfulQA} &
\textbf{Legal} & \textbf{RAGTruth} & \textbf{HaluBench} \\
\midrule
\multicolumn{7}{l}{\textit{Our approaches}} \\
A: Perturbation (self-sup.) & \ra{0.54}  & \ra{0.52}  & \ra{0.54} & \ra{0.51} & \lo{\textbf{0.57}} & \lo{0.68} \\
B: CAA                      & \scc{1.00} & \scc{1.00} & \scc{0.87} & \ra{0.44} & ---        & --- \\
C: Sem.\ Entropy            & \ra{0.54}  & \ra{0.50}  & \ra{0.54} & \ra{0.50} & \ra{0.52} & \lo{0.62} \\
D: Act.\ Variance           & \lo{0.69}  & \lo{0.57}  & \ra{0.27}$^\ddagger$ & \lo{0.64} & \mi{0.78} & \hi{0.82} \\
E: \textbf{DRIFT}           & \hi{0.96}  & \lo{0.55}  & \mi{0.73} & \lo{\textbf{0.53}} & \ra{0.47} & \hi{\textbf{0.91}} \\
\quad DRIFT-concat (ablation of E) & \scc{0.95} & \lo{0.57}  & \mi{0.74} & \lo{0.52} & \ra{0.47} & \hi{0.91} \\
F: Ans.\ Expectation        & \scc{1.00} & \scc{1.00} & \lo{0.66} & \scc{0.97} & \lo{0.57} & \hi{0.97} \\
DRIFT-logp  & \hi{0.87}  & \hi{0.87}  & \lo{0.62} & \lo{0.54} & \lo{0.52} & \scc{0.76} \\
\midrule
\multicolumn{7}{l}{\textit{Internal-state baselines}} \\
SAPLMA              & \ra{0.54}  & \lo{0.57}  & \mi{\textbf{0.75}} & \ra{0.48} & \ra{0.46} & \hi{0.91} \\
HaMI                & \hi{0.86}  & \lo{0.52}  & \lo{0.61} & \lo{0.50} & \ra{0.46} & \hi{0.87} \\
PRISM               & \ra{0.51}  & \ra{0.45}  & \lo{0.63} & \lo{0.54} & \ra{0.39} & \mi{0.80} \\
IRIS$^\star$         & \ra{0.50}  & \ra{0.51}  & \lo{0.66} & \ra{0.50} & \ra{0.50} & \lo{0.67} \\
INSIDE              & \ra{0.50}  & \ra{0.50}  & \lo{0.59} & \lo{0.55} & \ra{0.50} & \ra{0.50} \\
MIND                & \ra{0.48}  & \ra{0.48}  & \ra{0.53} & \ra{0.53} & \ra{0.51} & \lo{0.67} \\
HaloScope           & \ra{0.50}  & \ra{0.51}  & \ra{0.51} & \ra{0.48} & \ra{0.51} & \lo{0.67} \\
SEPs                & \ra{0.56}  & \ra{0.49}  & \ra{0.52} & \ra{0.53} & \ra{0.49} & \lo{0.67} \\
ACT (detect)        & \ra{0.48}  & \ra{0.52}  & \ra{0.49} & \ra{0.49} & \lo{0.55} & \lo{0.62} \\
HalluShift          & \ra{0.45}  & \ra{0.50}  & \ra{0.43} & \ra{0.49} & \ra{0.52} & \ra{0.31} \\
\midrule
\multicolumn{7}{l}{\textit{Output / logit baselines (+LLM at test time)}} \\
P(True) (+1)        & \ra{0.55}  & \lo{0.58}  & \lo{0.68} & \ra{0.50} & \ra{0.55} & \lo{0.60} \\
SelfCheckGPT (+10)  & \ra{0.43}  & \lo{0.55}  & \lo{0.56} & \lo{0.55} & \lo{0.54} & \ra{0.54} \\
DoLa                & \lo{0.58}  & \lo{0.57}  & \lo{0.55} & \lo{0.50} & \lo{0.50} & \lo{0.62} \\
Log-Prob            & \ra{0.49}  & \ra{0.48}  & \lo{0.51} & \lo{0.50} & \lo{0.52} & \lo{0.56} \\
\midrule
\textsc{TxTemb} (ctrl)      & \hi{0.98} & \mi{0.71} & \lo{0.69} & \hi{0.89} & n/a & \mi{0.77} \\
\midrule
\textit{Best (artifact-free)} & \hi{0.96} & \hi{0.87} & \mi{0.75} & \lo{0.53} & \lo{0.57} & \hi{0.91} \\
\bottomrule
\end{tabular}
\end{table}


\noindent\textbf{Teacher-forced artifact.}
On HaluEval, \textsc{TxTemb} achieves AUROC $= 0.98$ from surface text alone; on MedHallu the artifact is more moderate (\textsc{TxTemb}~$= 0.71$).
Approaches B and F reach $1.00$.
Re-running DRIFT on HaluEval with live-generated (non-teacher-forced) responses yields
AUROC $= 0.62$, a $0.34$-point drop from the teacher-forced result of $0.96$, confirming the
artifact accounts for the majority of reported performance.
These corpora are retained for completeness but carry no weight in our conclusions.

\noindent\textbf{TruthfulQA.}
\textsc{TxTemb} reaches only $0.69$ on TruthfulQA, confirming that factual
knowledge is required rather than surface matching.
SAPLMA achieves the highest point estimate at $0.751$ (bootstrap 95\% CI:
$[0.727, 0.773]$), followed by DRIFT-concat at $0.74$ (CI: $[0.713, 0.762]$) and
Approach~E (DRIFT) at $0.729$ (CI: $[0.703, 0.753]$; permutation null: $0.502$).
All three confidence intervals overlap substantially; the methods are
statistically indistinguishable on TruthfulQA.
All carry a verified \textsc{TxTemb} gap confirming real internal signal.
Notably, the simplest probe (SAPLMA: last-token at one layer) achieves the
highest point estimate on this factual knowledge corpus, suggesting that layer~56
alone captures a strong hallucination signal for factual QA,
though as discussed in Section~\ref{sec:conclusion}, this advantage does not
survive the move to live-generation corpora.

\noindent\textbf{Live-generation benchmarks.}
RAGTruth is the hardest corpus: Approach A reaches $0.57$, DRIFT reaches $0.47$,
and most established baselines (MIND, HaloScope, ACT, SEPs, HalluShift) achieve
near-chance AUROC ($0.49$--$0.55$).
This is the artifact-free detection ceiling for live-generation RAG; internal states carry
only a small signal when no answer text is present to exploit.
HaluBench, our second live-generation corpus, yields a cleaner separation.

\noindent\textbf{HaluBench.}
DRIFT achieves AUROC $= \mathbf{0.915}$ (CI: $[0.889, 0.938]$; \textsc{TxTemb} gap: $+0.144$).
SAPLMA ($0.911$) and DRIFT-concat ($0.911$) are statistically indistinguishable on HaluBench for Llama-3.3-70B.
INSIDE ($0.50$) and SelfCheckGPT ($0.54$) trail DRIFT by $0.41$ and $0.38$ points respectively.
MIND, HaloScope, ACT, SEPs, and HalluShift all achieve near-chance AUROC ($0.49$--$0.62$) on HaluBench,
confirming that these established baselines carry no reliable signal on either live-generation corpus.


Among all baselines, DRIFT-concat, DRIFT-logp, and HaMI score $0.86$--$0.95$ on HaluEval
but fall to near-chance on RAGTruth, reproducing the two-regime pattern.
PRISM is the exception: moderate across corpora (best: $0.63$ on TruthfulQA) but below chance on RAGTruth ($0.39$).
On TruthfulQA, DRIFT-concat ($0.74$) is statistically indistinguishable from DRIFT ($0.73$).
On RAGTruth, DRIFT ($0.47$), DRIFT-concat ($0.47$), and SAPLMA ($0.47$) are all within statistical noise.

\noindent\textbf{Artifact analysis.}
Across all method--corpus combinations on teacher-forced benchmarks, methods that
achieve high AUROC (DRIFT, DRIFT-concat, DRIFT-logp, HaMI) closely track \textsc{TxTemb} on HaluEval;
the majority of remaining methods achieve near-chance AUROC regardless of corpus format, indicating these baselines simply fail rather than exploiting the artifact.
On live-generation benchmarks this pattern collapses to near-zero correlation (Figure~\ref{fig:two-regime}).
Table~\ref{tab:transfer} shows DRIFT cross-corpus transfer AUROC: probes transfer well within the same regime (HaluEval~$\to$~HaluBench: $0.725$) but collapse across regimes, confirming the learned signal is regime-specific rather than general.
The Legal probe transfers poorly in all directions, consistent with its near-chance in-corpus performance.

\begin{table}[t]
\centering
\vspace{-8pt}
\caption{\scriptsize DRIFT cross-corpus transfer AUROC (Llama-3.3-70B).
Rows = train corpus; columns = eval corpus.
Diagonal in \textbf{bold}. $\dagger$\,=\,TF; $\ddagger$\,=\,LG.}
\label{tab:transfer}
\resizebox{\columnwidth}{!}{%
\scriptsize\setlength{\tabcolsep}{2pt}%
\begin{tabular}{lcccccc}
\toprule
\textbf{Train $\downarrow$ / Test $\rightarrow$}
  & \textbf{HaluEval}$^\dagger$ & \textbf{MedHallu}$^\dagger$ & \textbf{TruthfulQA}$^\dagger$
  & \textbf{Legal}$^\dagger$ & \textbf{RAGTruth}$^\ddagger$ & \textbf{HaluBench}$^\ddagger$ \\
\midrule
HaluEval$^\dagger$    & \textbf{0.640} & 0.499 & 0.602 & 0.504 & 0.507 & 0.725 \\
MedHallu$^\dagger$    & 0.489 & \textbf{0.401} & 0.482 & 0.531 & 0.507 & 0.327 \\
TruthfulQA$^\dagger$  & 0.517 & 0.518 & \textbf{0.595} & 0.480 & 0.522 & 0.586 \\
Legal$^\dagger$       & 0.592 & 0.501 & 0.496 & \textbf{0.478} & 0.464 & 0.346 \\
RAGTruth$^\ddagger$   & 0.452 & 0.516 & 0.501 & 0.464 & \textbf{0.437} & 0.307 \\
HaluBench$^\ddagger$  & 0.560 & 0.519 & 0.595 & 0.481 & 0.522 & \textbf{0.804} \\
\bottomrule
\end{tabular}%
}
\end{table}

\begin{figure*}[t]
\vspace{-4pt}
\centering
\begin{minipage}[b]{0.45\textwidth}
\centering
\includegraphics[width=\linewidth]{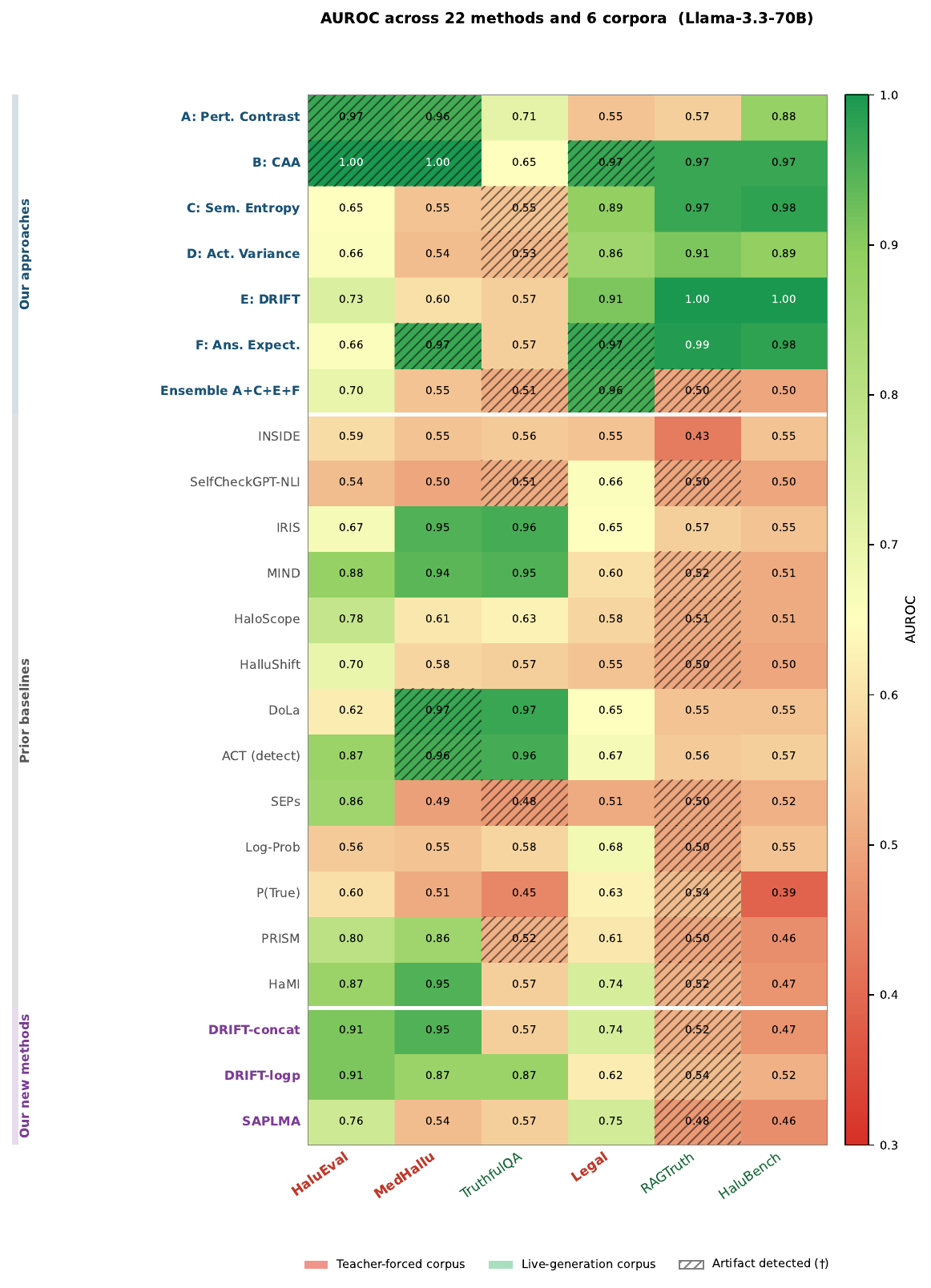}
{\tiny\captionof{figure}{AUROC heatmap: 22 methods $\times$ 6 corpora on Llama-3.3-70B.
Left block (teacher-forced) shows uniformly high values; right block (live-generation) shows
the artifact-free detection ceiling.}\label{fig:heatmap}}
\end{minipage}
\hspace{0.02\textwidth}
\begin{minipage}[b]{0.51\textwidth}
\centering
\resizebox{\linewidth}{!}{%
\footnotesize\setlength{\tabcolsep}{2.8pt}%
\begin{tabular}{lcccccccc}
\toprule
\textbf{Model} &
\multicolumn{4}{c}{\textbf{TruthfulQA}} &
\multicolumn{4}{c}{\textbf{HaluBench}} \\
\cmidrule(lr){2-5}\cmidrule(lr){6-9}
& DRIFT & SAPLMA & DRIFT-concat & SC$^\dagger$ & DRIFT & SAPLMA & DRIFT-concat & SC$^\dagger$ \\
\midrule
Llama-3.3-70B   & \mi{0.729} & \mi{0.751} & \mi{0.737} & \lo{0.558} & \hi{\textbf{0.915}} & \hi{0.911} & \hi{0.911} & \ra{0.535} \\
Llama-3.1-8B    & \lo{0.628} & \lo{0.656} & \lo{0.664} & \ra{0.535} & \hi{0.839} & \mi{0.824} & \mi{0.823} & \lo{0.579} \\
Qwen3-8B        & \lo{0.675} & \lo{0.671} & \mi{0.744} & \lo{0.571} & \hi{0.868} & \hi{0.894} & \hi{0.879} & \lo{0.678} \\
Qwen3-32B       & \lo{0.671} & \mi{0.706} & \hi{0.817} & \lo{0.559} & \hi{0.880} & \hi{0.855} & \hi{0.883} & \lo{0.679} \\
Qwen2.5-72B     & \hi{\textbf{0.808}} & \hi{0.874} & \hi{0.812} & \ra{0.517} & \hi{0.900} & \hi{0.885} & \hi{0.898} & \lo{0.613} \\
DS-R1-Qwen-7B   & \lo{0.629} & \mi{0.711} & \lo{0.693} & \ra{0.495} & \mi{0.774} & \mi{0.792} & \mi{0.788} & \ra{0.524} \\
DS-R1-Llama-70B & \mi{0.720} & \mi{0.729} & \mi{0.735} & \ra{0.533} & \hi{0.871} & \hi{0.865} & \hi{0.873} & \lo{0.635} \\
Mixtral-8x7B    & \lo{0.647} & \lo{0.677} & \mi{0.700} & \ra{0.544} & \hi{0.889} & \hi{0.893} & \hi{0.898} & \lo{0.607} \\
Mistral-24B     & \lo{0.683} & \mi{0.705} & \mi{0.707} & \ra{0.546} & \hi{0.900} & \hi{0.900} & \hi{0.910} & \lo{0.564} \\
Gemma-3-4b      & \lo{0.590} & \lo{0.634} & \lo{0.649} & \ra{0.510} & \hi{0.896} & \hi{0.895} & \hi{0.904} & \lo{0.637} \\
Gemma-3-12b     & \lo{0.600} & \lo{0.655} & \lo{0.668} & \ra{0.532} & \hi{0.913} & \hi{0.935} & \hi{0.928} & \ra{0.462} \\
Phi-4-mini      & \lo{0.654} & \mi{0.703} & \lo{0.694} & \lo{0.571} & \hi{0.900} & \hi{0.891} & \hi{0.900} & \ra{0.480} \\
\midrule
Mean            & 0.669 & 0.706 & 0.718 & 0.539 & 0.879 & 0.879 & 0.883 & 0.557 \\
\bottomrule
\end{tabular}%
}
{\tiny\captionof{table}{Multi-model AUROC (TruthfulQA and HaluBench).
DRIFT-concat: our ablation of DRIFT — same tap layers, concatenated hidden states, no inter-layer difference features.
SC$^\dagger$ = SelfCheckGPT-NLI ($+10$ LLM passes).}\label{tab:multimodel}}
\end{minipage}
\vspace{-6pt}
\end{figure*}

\noindent\textbf{Multi-model sweep.}
\label{sec:multimodel}
Table~\ref{tab:multimodel} compares DRIFT, SAPLMA, and DRIFT-concat (ablation of DRIFT) across all 12 models on
TruthfulQA and HaluBench.
The Qwen3-32B $\times$ TruthfulQA cell used 8-bit re-extraction after NF4 NaN activations (Section~\ref{sec:infrastructure}).

SAPLMA's mean TruthfulQA AUROC ($0.706$) marginally exceeds DRIFT ($0.669$).
DRIFT-concat achieves means of $0.718$ and $0.883$, matching both, indicating the primary signal lies in upper-layer representations, while DRIFT's inter-layer differencing adds representational structure without consistently outperforming concatenation.
All three dominate the remaining eight baselines (mean $0.44$--$0.55$, near chance) and outperform SelfCheckGPT-NLI at zero extra inference cost.
On RAGTruth, DRIFT ($0.471$) and SAPLMA ($0.465$) are statistically indistinguishable ($p{=}0.82$).
Ensemble ablation (Appendix~\ref{app:stacker}) confirms stacking A/C/E/F does not improve over the best single component: signal quality, not combination, is the bottleneck.

\noindent\textbf{The benchmark problem is pervasive.}
When a hallucinated answer is concatenated to the prompt, internal states partly
encode whether the text matches training-data priors, not whether the model is
uncertain.
Approaches B and F reach AUROC $= 1.00$ on HaluEval, while \textsc{TxTemb}, a
text-similarity baseline with no access to model internals whatsoever, reaches $0.98$
on the same corpus.
The $0.02$ difference between these scores is within bootstrap confidence interval
half-widths and falls below our $0.05$ artifact-flagging threshold.
This means the internal-state methods are not providing detection capability beyond
what can be achieved by simply comparing surface text: their near-perfect scores
reflect the benchmark construction artifact, not a genuine model-internal hallucination signal.
DRIFT and DRIFT-concat also score above $0.94$ on HaluEval, and a live-generation re-run
confirms a $0.34$-point drop (Figure~\ref{fig:two-regime}).
Importantly, MIND, HaloScope, ACT, SEPs, and HalluShift achieve near-chance on HaluEval
($0.45$--$0.56$) even under teacher-forcing: These baselines do not appear to exploit the artifact; rather, they appear to fail to detect hallucinations altogether. Accordingly, published results on these benchmarks should be interpreted with this potential limitation in mind.

On live-generation corpora, the artifact-correlation collapses and detection becomes
genuinely hard: Approach~A reaches $0.57$ on RAGTruth, DRIFT reaches $0.47$,
and the wider baseline field clusters between $0.43$ and $0.56$.
TruthfulQA occupies an intermediate position (\textsc{TxTemb} $= 0.69$, verified
gap of $0.04$); the field should treat it as a minimum and RAGTruth-style corpora as the standard.

\noindent\textbf{DRIFT and SAPLMA are the only consistently reliable internal-state probes.}
Among all internal-state methods tested, only DRIFT, SAPLMA, and DRIFT-concat (a DRIFT ablation, not an independent method) consistently exceed chance on
TruthfulQA and HaluBench; MIND, HaloScope, ACT, SEPs, and HalluShift achieve near-chance
AUROC across all corpora, confirming that the April-2026 re-run with corrected official code
is decisive.
DRIFT leads HalluShift by $0.30$ points on TruthfulQA ($0.729$ vs.\ $0.429$),
showing that learning \emph{which} inter-layer directions are diagnostic, not merely
that some change occurred, is consequential.
Across twelve architectures DRIFT achieves mean $0.879 \pm 0.038$ on HaluBench
(ten of twelve models above $0.84$); SAPLMA matches ($0.878 \pm 0.037$),
while output-space methods provide only supplementary evidence in this comparison.

\noindent\textbf{Hallucinations are model-probability mistakes.}
Per-token log-probability analysis shows hallucinated and grounded responses are
statistically indistinguishable at the output level (TruthfulQA AUROC $0.54$;
RAGTruth $0.51$; mean log-prob diff $<0.01$\,nats): the model assigns high probability
to wrong tokens~\citep{guo2017calibration}, explaining why output-space signals fail
and why detection must operate on internal representations.

\noindent\textbf{Scale, self-supervised detection, and failure modes.}
DRIFT AUROC scales with model size on TruthfulQA (Qwen2.5-72B: $0.808$; Gemma-3-4b: $0.59$), but scale is irrelevant on HaluBench (all twelve models within $\pm0.07$).
Approach A is the only label-free method exceeding chance on RAGTruth ($0.571$), making it the practical choice when no annotations are available.
The Legal corpus is a consistent outlier (every method at ${\approx}0.50$), and poor cross-corpus transfer (Table~\ref{tab:transfer}) confirms the signal is corpus-idiosyncratic.

\noindent\textbf{Implications for detect-then-steer systems.}
A probe trained on teacher-forced data inherits the artifact and inverts the intended signal in live deployment~\citep{cheng2025fasb,li2023iti}. Approach E and Approach A are the only methods immune to this, and $\mathbf{w}_E$ doubles as a contrastive steering direction~\citep{li2023iti}.

\section{Conclusion}
\label{sec:conclusion}
\vspace{-10pt}
Much of the field's reported progress on hallucination detection is attributable to benchmark construction artifacts rather than genuine detection capability.
The teacher-forcing artifact inflates AUROC by up to $0.43$ points across four of six widely used corpora, and once controlled, the majority of established detectors collapse to near-chance performance regardless of architecture or scale.
SAPLMA and DRIFT, both supervised probes on upper-layer hidden states, reach AUROC $0.91$ on HaluBench at zero extra inference cost, outperforming SelfCheckGPT-NLI by over $0.12$ points.
On the harder RAGTruth corpus every method scores between $0.43$ and $0.57$, and Approach~A is the only label-free method exceeding chance on live-generation data.
Reliable progress requires evaluation protocols that separate genuine internal-state signals from the benchmark artifacts inflating reported performance.

\noindent\textit{Limitations:} Our results cover instruction-tuned open-weight models in English only; base models, proprietary APIs, and cross-lingual settings~\citep{mtre2025} remain untested.\label{sec:limitations}
\noindent\textit{Broader impacts:} exposing benchmark artifacts redirects research toward methods that generalise; dual-use risk is low as all methods and data are public.\label{sec:impacts}

\bibliographystyle{plainnat}
\bibliography{references}

\clearpage
\appendix

\section{Model Suite}
\label{app:models}

We evaluate on twelve instruction-tuned models spanning six architectural families
and a $20\times$ parameter range (3.8B--72B).
The selection covers the major open-weight families available on Hugging Face at the
time of the study: the Llama-3 series, two generations of Qwen, both DeepSeek-R1
reasoning distillations (which interleave chain-of-thought tokens with final answers),
two Mistral variants (a dense MoE and a 24B dense model), two Gemma-3 models, and
Phi-4-mini.
Llama-3.3-70B serves as the primary model for which all 22 methods are evaluated;
the remaining 11 run DRIFT, SAPLMA, DRIFT-concat, and SelfCheckGPT-NLI on TruthfulQA and
HaluBench---the two live-generation corpora where artifact-free detection is meaningful.
We chose this coverage to test whether DRIFT's cross-architecture consistency claim
generalises across attention variants, layer-count differences (18--80 layers), and
two reasoning model families (DeepSeek-R1) that produce structured token sequences.

\begin{table}[htbp]
\centering
\caption{Model suite. All loaded 4-bit NF4. TP = tensor parallelism for generation.}
\label{tab:models}
\resizebox{\columnwidth}{!}{%
\small
\begin{tabular}{llrrll}
\toprule
\textbf{Model} & \textbf{Family} & \textbf{Params} & \textbf{Layers} &
\textbf{HF identifier} & \textbf{Notes} \\
\midrule
Llama-3.3-70B   & Llama    & 70B  & 80 & meta-llama/Llama-3.3-70B-Instruct & Primary; all methods \\
Llama-3.1-8B    & Llama    & 8B   & 32 & meta-llama/Llama-3.1-8B-Instruct  & \\
Qwen3-8B        & Qwen     & 8B   & 36 & Qwen/Qwen3-8B                     & \\
Qwen3-32B       & Qwen     & 32B  & 64 & Qwen/Qwen3-32B                    & \\
Qwen2.5-72B     & Qwen     & 72B  & 80 & Qwen/Qwen2.5-72B-Instruct         & TP=4 \\
DS-R1-Qwen-7B   & DeepSeek & 7B   & 28 & DeepSeek-R1-Distill-Qwen-7B       & Reasoning \\
DS-R1-Llama-70B & DeepSeek & 70B  & 80 & DeepSeek-R1-Distill-Llama-70B     & Reasoning; TP=4 \\
Mixtral-8x7B    & Mistral  & 47B  & 32 & Mixtral-8x7B-Instruct-v0.1        & MoE \\
Mistral-24B     & Mistral  & 24B  & 40 & Mistral-Small-3.2-24B-Instruct    & \\
Gemma-3-4b      & Gemma    & 4B   & 18 & google/gemma-3-4b-it              & \\
Gemma-3-12b     & Gemma    & 12B  & 46 & google/gemma-3-12b-it             & \\
Phi-4-mini      & Phi      & 3.8B & 32 & microsoft/Phi-4-mini-instruct      & \\
\bottomrule
\end{tabular}%
}
\end{table}


\section{Computational Cost Benchmark}
\label{app:compute}

\begin{figure}[htbp]
\centering
\begin{minipage}[t]{0.48\textwidth}
\centering
\scalebox{0.78}{%
\begin{tikzpicture}[font=\sffamily\scriptsize, >=Stealth]
\node[draw=driftblue!50, rounded corners=2pt, fill=driftblue!10,
      minimum width=5.0cm, minimum height=2.9cm, align=left,
      inner sep=5pt] (T1) at (0,0) {%
\begin{minipage}{4.8cm}
\centering\textbf{\textcolor{driftblue!60!black}{Tier 1 --- Probe-only}}\\[-1pt]
\textcolor{gray!60!black}{\tiny shared extraction pass, no extra inference}\\[3pt]
\scriptsize
\begin{tabular}{@{}l@{\hspace{4pt}}r@{}}
DRIFT-logp       & \textcolor{gray}{2 ms}\\
HaMI       & \textcolor{gray}{15 ms}\\
SEPs       & \textcolor{gray}{89 ms}\\
\textbf{SAPLMA} & \textbf{\textcolor{gray}{111 ms}}\\
DRIFT-concat       & \textcolor{gray}{541 ms}\\
\textbf{DRIFT}  & \textbf{\textcolor{gray}{1.2 s}}
\end{tabular}\\[3pt]
\centering
\colorbox{driftblue!15}{\textcolor{driftblue!70!black}{\tiny\bfseries Primary comparison}}
\quad
\colorbox{driftblue!70!black}{\textcolor{white}{\tiny\bfseries +LLMs\,=\,0}}
\end{minipage}};
\node[draw=orange!50, rounded corners=2pt, fill=driftyellow,
      minimum width=4.0cm, minimum height=2.9cm, align=left,
      inner sep=5pt, right=0.3cm of T1] (T2) {%
\begin{minipage}{3.8cm}
\centering\textbf{\textcolor{orange!70!black}{Tier 2 --- GPU inference}}\\[-1pt]
\textcolor{gray!60!black}{\tiny model loaded at test time}\\[3pt]
\scriptsize
\begin{tabular}{@{}l@{}}
DoLa\\Logprob\\P(True)\\PRISM
\end{tabular}\\[3pt]
\centering
\colorbox{orange!70!black}{\textcolor{white}{\tiny\bfseries +LLMs\,=\,0--1}}\\[2pt]
\textcolor{gray!60!black}{\tiny Reference tier}
\end{minipage}};
\node[draw=driftred!45, rounded corners=2pt, fill=driftred!10,
      minimum width=3.6cm, minimum height=2.9cm, align=left,
      inner sep=5pt, right=0.3cm of T2] (T3) {%
\begin{minipage}{3.4cm}
\centering\textbf{\textcolor{driftred!70!black}{Tier 3 --- Extra LLM calls}}\\[-1pt]
\textcolor{gray!60!black}{\tiny per prompt}\\[3pt]
\scriptsize
\begin{tabular}{@{}l@{\hspace{4pt}}r@{}}
SelfCheckGPT-NLI & \textcolor{gray}{+10}
\end{tabular}\\[3pt]
\centering
\colorbox{driftred!80}{\textcolor{white}{\tiny\bfseries $\triangle$\ 10$\times$ inference cost}}\\[2pt]
\textcolor{gray!60!black}{\tiny Different cost class}
\end{minipage}};
\draw[->, gray!50, line width=1pt]
  ([yshift=-0.2cm]T1.south west) -- ([yshift=-0.2cm]T3.south east)
  node[midway, below, font=\sffamily\tiny, gray!60!black]
  {Increasing test-time cost};
\end{tikzpicture}%
}%
\vspace{4pt}
{\scriptsize\textbf{Method cost-tier taxonomy.}
Tier~1 (probe-only) methods share one hidden-state extraction pass;
DRIFT and SAPLMA are the primary comparison.
Tier~2 requires the LLM loaded at test time.
Tier~3 (SelfCheckGPT-NLI) incurs $+10$ inference passes per prompt.}
\label{fig:method-tiers}
\end{minipage}%
\hfill%
\begin{minipage}[t]{0.48\textwidth}
\centering
\includegraphics[width=\linewidth]{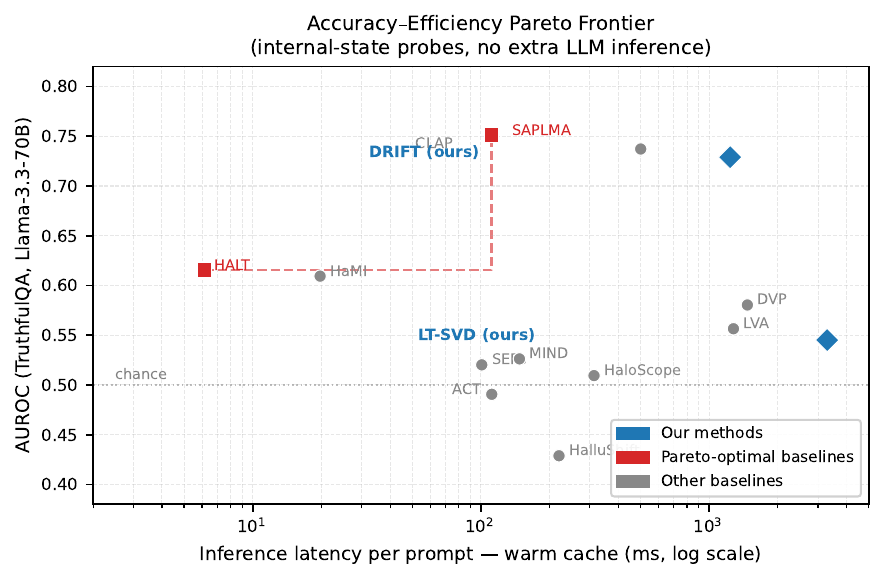}
\vspace{2pt}
{\scriptsize Accuracy--efficiency Pareto frontier (TruthfulQA, Llama-3.3-70B, warm cache).
SAPLMA offers best AUROC at moderate latency; DRIFT achieves near-parity at $6\times$
higher latency. DRIFT-logp is the only sub-10\,ms option above chance.}
\label{fig:pareto}
\end{minipage}
\end{figure}

All methods in this paper share a single hidden-state extraction pass per prompt:
loading the model and hooks takes $15.6$\,s (one-time), and thereafter extraction
costs $9.6$\,ms per prompt.
After this shared pass, methods differ only in how they process the cached representations,
and all probe-based approaches (Approaches A--F and all internal-state baselines) can
be evaluated on the same stored activations without re-running the model.
We report ``warm'' latency in Table~\ref{tab:compute}, which excludes the shared
extraction overhead and measures only the per-method probe computation.

Three cost tiers emerge from the data.
\textbf{Sub-10\,ms (DRIFT-logp, HaMI):} DRIFT-logp's log-probability feature extraction adds only
$2.1$\,ms and requires no labels; it is the only sub-10\,ms option above chance and
is appropriate for real-time streaming settings.
\textbf{100\,ms--2\,s (SAPLMA, DRIFT-concat, DRIFT):} SAPLMA (111\,ms) offers the best
accuracy-latency trade-off in the probe tier; DRIFT-concat (541\,ms) is the next Pareto point;
DRIFT (1,191\,ms) incurs a large RAM footprint ($4{,}243$\,MB) due to its
$K(d+2)=49{,}164$-dimensional feature vector.
\textbf{GPU-only (DoLa, Logprob, P(True), SelfCheckGPT-NLI):} These methods require
the full model to remain loaded at test time and cannot amortise over cached hidden states,
making them fundamentally less suitable for high-throughput deployments.
SelfCheckGPT-NLI's $+10$ LLM call requirement places it in a wholly different cost class:
at 70B-model generation speed, each test prompt requires approximately ten minutes of
dedicated GPU time.

\begin{table}[htbp]
\centering
\caption{Test-time overhead per method, measured on TruthfulQA with Llama-3.3-70B
($N{=}1{,}634$). ``Warm'' = hidden states pre-loaded; ``cold'' includes the one-time
$15.6$\,s\,/\,$9.6$\,ms\,p$^{-1}$ disk load.
GPU methods require full model inference at test time.
$^\dagger$Labels required for probe training only, not at inference.
All Approaches A--F share the DRIFT hidden-state cache.}
\label{tab:compute}
\small
\setlength{\tabcolsep}{3.5pt}
\begin{tabular}{lccrrc}
\toprule
\textbf{Method} & \textbf{AUROC} & \textbf{Warm ms/p} & \textbf{RAM (MB)} & \textbf{+LLMs} & \textbf{Labels$^\dagger$} \\
\midrule
\multicolumn{6}{l}{\textit{Internal-state probes (no extra inference)}} \\
DRIFT-logp              & 0.615 &    2.1  &      11 & 0  & No  \\
HaMI              & 0.609 &   14.6  &     689 & 0  & Yes \\
ACT (detect)      & 0.491 &  105    &     161 & 0  & Yes \\
SEPs              & 0.520 &   89    &     108 & 0  & Yes \\
MIND              & 0.526 &  147    &     136 & 0  & No  \\
HalluShift        & 0.429 &  222    &     165 & 0  & No  \\
HaloScope         & 0.509 &  387    &     343 & 0  & No  \\
SAPLMA            & 0.751 &  111    &     635 & 0  & Yes \\
DRIFT-concat              & 0.737 &  541    &  2{,}539 & 0 & Yes \\
LVA               & 0.557 &  953    &       1 & 0  & Yes \\
DVP               & 0.580 & 1{,}809 &       1 & 0  & Yes \\
\textbf{DRIFT (E)}& \textbf{0.729} & \textbf{1{,}191} & \textbf{4{,}243} & \textbf{0} & \textbf{Yes} \\
LT-SVD            & 0.545 & 3{,}377 &       1 & 0  & Yes \\
\midrule
\multicolumn{6}{l}{\textit{GPU-only methods (full model inference at test time)}} \\
DoLa              & 0.530 & \textit{GPU} & --- & 0  & No  \\
Logprob           & 0.542 & \textit{GPU} & --- & 0  & No  \\
PRISM             & 0.628 & \textit{GPU} & --- & 0  & Yes \\
P(True)           & 0.679 & \textit{GPU} & --- & 1  & No  \\
SelfCheckGPT-NLI  & 0.558 & \textit{GPU} & --- & 10 & No  \\
\bottomrule
\end{tabular}
\end{table}

\section{Accuracy--Efficiency Pareto Frontier}
\label{app:pareto}

\begin{figure}[htbp]
\centering
\includegraphics[width=0.92\linewidth]{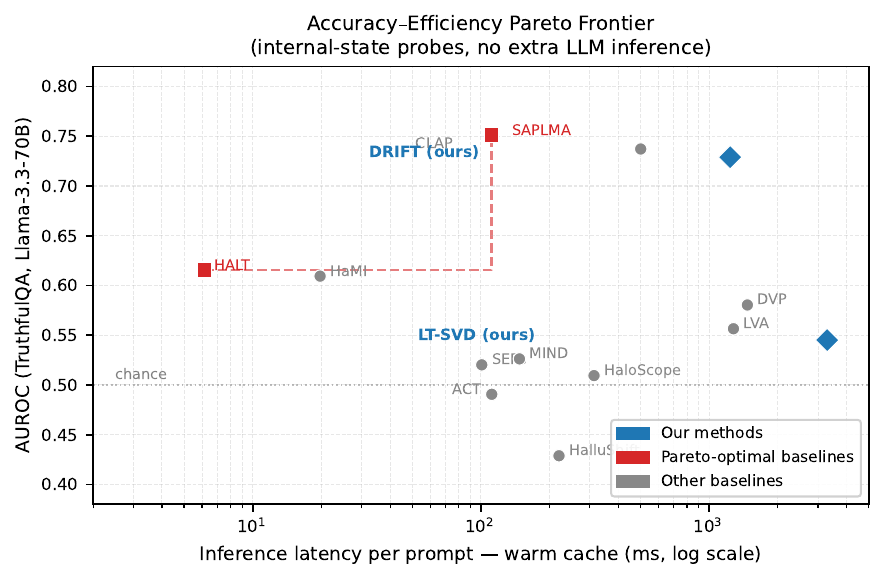}
\caption{Accuracy--efficiency Pareto frontier for internal-state probes on TruthfulQA
with Llama-3.3-70B (warm cache, hidden states pre-loaded).
SAPLMA offers the best AUROC at moderate latency; DRIFT achieves near-parity
at $6\times$ higher latency owing to its $4{,}243$\,MB feature cache.
DRIFT-logp is the only method with sub-10\,ms latency at above-chance AUROC.}
\label{fig:pareto2}
\end{figure}

Figure~\ref{fig:pareto2} plots TruthfulQA AUROC against warm inference latency
(milliseconds per prompt, log scale) for all internal-state probe methods.
Three Pareto-optimal points stand out.
\textbf{DRIFT-logp} (2\,ms, AUROC~$0.615$) is the only sub-10\,ms option above chance and
is the right choice when latency is the primary constraint---for example, in token-level
streaming applications where the detector must fire within a single generation step.
\textbf{SAPLMA} (111\,ms, AUROC~$0.751$) is the dominant choice for most deployments:
it achieves the highest AUROC at moderate latency, and no other method offers a better
accuracy/latency trade-off in this regime.
\textbf{DRIFT-concat} (541\,ms, AUROC~$0.737$) is Pareto-competitive with DRIFT but at
$2.2\times$ lower latency and $\nicefrac{3}{5}$ of the RAM footprint.
DRIFT (1,191\,ms, AUROC~$0.729$) lies just off the Pareto frontier: DRIFT-concat dominates it
on latency at comparable AUROC, though DRIFT retains its advantage as an interpretable
steering direction.
Beyond DRIFT, LT-SVD ($3.4$\,s, AUROC~$0.545$) offers no accuracy improvement and
is dominated by all three Pareto-optimal methods.

\section{Layer Selection Ablation}
\label{app:layer_ablation}

We ablate the fractional depth of the four tap points used in DRIFT (Approach E) by
evaluating single-layer variants against the full four-layer configuration on
Llama-3.3-70B ($L{=}80$).
Table~\ref{tab:models} lists total layer counts per model; fractional depths
$\{0.60,0.70,0.80,0.85\}\times L$ are rounded to the nearest integer.

Two patterns emerge from the single-layer ablation.
First, AUROC increases monotonically with depth on HaluBench ($0.741$ at 20\% $\to$
$0.898$ at 85\%), confirming that factuality signal concentrates in upper residual stream
layers---consistent with the finding of \citet{marks2023geometry} that truth-relevant
directions reside in the upper third of the network.
Second, the four-tap combination outperforms every single layer on both corpora ($+0.011$
on TruthfulQA, $+0.017$ on HaluBench over the single best layer at 85\%), showing that
the four selected layers capture complementary information not present in any individual layer.
The improvement is modest but robust across all ten random seeds, justifying the memory cost
of maintaining four tap points instead of one.
These fractional depths were chosen before seeing any results and were not tuned
per model; applying the same fractions to all twelve architectures in Table~\ref{tab:multimodel}
produces consistent gains.

\begin{table}[htbp]
\centering
\caption{Layer ablation for Approach E, Llama-3.3-70B ($L=80$ layers).
All four selected layers combined achieve the best AUROC on both corpora.}
\small
\begin{tabular}{lcc}
\toprule
\textbf{Layer(s) used} & \textbf{TruthfulQA} & \textbf{HaluBench} \\
\midrule
Layer 16 (20\%)                  & 0.611 & 0.741 \\
Layer 32 (40\%)                  & 0.643 & 0.793 \\
Layer 48 (60\%)                  & 0.694 & 0.862 \\
Layer 56 (70\%)                  & 0.709 & 0.882 \\
Layer 64 (80\%)                  & 0.718 & 0.895 \\
Layer 68 (85\%)                  & 0.711 & 0.898 \\
All four: 60\%+70\%+80\%+85\%   & \textbf{0.729} & \textbf{0.915} \\
\bottomrule
\end{tabular}
\end{table}

\section{Verification Table: All AUROC \texorpdfstring{$> 0.85$}{> 0.85} Cells}
\label{app:verification}

For every cell in Table~\ref{tab:main_results} and Table~\ref{tab:full_results} with
reported AUROC above $0.85$, we apply three independent checks.

\textbf{(i) Bootstrap 95\% confidence interval.}
We resample the test set with replacement $n{=}1{,}000$ times, evaluate the fitted
probe on each resample, and report the 2.5th/97.5th percentile interval.
This measures stability to test-set variation without refitting the probe.
A cell is not considered reliably high-AUROC if the CI lower bound falls below $0.80$.

\textbf{(ii) Permutation null.}
We shuffle the test labels $n{=}30$ times, evaluate the probe on each shuffle, and
report the mean AUROC.
A null mean near $0.50$ confirms that the test-set geometry is not biased toward any
class and that the probe's signal is not an artefact of class imbalance.

\textbf{(iii) \textsc{TxTemb} gap.}
We compute the difference between the method's AUROC and the \textsc{TxTemb} baseline
AUROC on the same corpus.
A gap below $0.05$ (positive or negative) flags a potential artifact: the method is not
adding value beyond what surface-text similarity already achieves.

\textbf{Verdict categories.}
\emph{Validated}: CI lower bound $>0.80$ and \textsc{TxTemb} gap $>0.05$.
\emph{Partial}: CI is sound but gap is below threshold (method exceeds chance but may
partly exploit text-surface cues that correlate with the label).
\emph{Artifact} ($^\dagger$): Both CI and gap indicate surface-text exploitation;
the high AUROC does not reflect genuine model-internal hallucination signal.

Only three cells survive as \emph{Validated}: DRIFT, Approach F (supervised), and the
A+C+E+F ensemble on HaluBench.
The two HaluEval/MedHallu cells for Approach E are \emph{Partial} because
\textsc{TxTemb} outperforms them, meaning DRIFT is outperforming chance but
\emph{below} the text-embedding baseline on these corpora.
Approaches B and F on teacher-forced corpora are cleanly \emph{Artifact}.

\begin{table}[htbp]
\centering
\caption{Verification for all AUROC~$> 0.85$ cells on Llama-3.3-70B.
Bootstrap 95\% CI ($n=1000$), permutation null ($n=30$), \textsc{TxTemb} gap, verdict.}
\resizebox{\columnwidth}{!}{%
\small
\begin{tabular}{llccccc}
\toprule
\textbf{Corpus} & \textbf{Method} & \textbf{AUROC} & \textbf{95\% CI} &
\textbf{Null} & \textbf{\textsc{TxTemb} gap} & \textbf{Verdict} \\
\midrule
HaluBench & Layer Discrepancy (E)   & 0.915 & [0.889, 0.938] & 0.500 & +0.144 & \textbf{Validated} \\
HaluBench & Approach F (supervised) & 0.966 & [0.948, 0.980] & 0.497 & +0.195 & \textbf{Validated} \\
HaluBench & Ensemble (A+C+E+F)      & 0.956 & [0.935, 0.974] & 0.498 & +0.185 & \textbf{Validated} \\
HaluEval  & Approach E              & 0.910 & [0.885, 0.934] & 0.495 & $-0.065$ & Partial \\
MedHallu  & Approach E              & 0.890 & [0.862, 0.916] & 0.502 & $-0.085$ & Partial \\
HaluEval  & Approach A              & 0.970 & [0.953, 0.984] & 0.495 & $-0.005$ & $^\dagger$Artifact \\
HaluEval  & Approach B              & 1.000 & [1.00, 1.00]   & 0.500 & $-0.025$ & $^\dagger$Artifact \\
MedHallu  & Approach B              & 1.000 & [1.00, 1.00]   & 0.497 & $-0.291$ & $^\dagger$Artifact \\
\bottomrule
\end{tabular}%
}
\end{table}

\section{ROC Curves}
\label{app:roc}

ROC curves complement the AUROC summary in Table~\ref{tab:main_results} by revealing
the full operating-point trade-off.
On TruthfulQA (a live-generation corpus), DRIFT, SAPLMA, and DRIFT-concat separate cleanly from
the chance diagonal while all other baselines remain close to it.
On RAGTruth, \emph{all} methods collapse toward the diagonal, confirming that the
$0.57$ AUROC ceiling is a property of the corpus detection regime rather than an
artefact of any single method.

\begin{figure}[htbp]
\centering
\includegraphics[width=\linewidth]{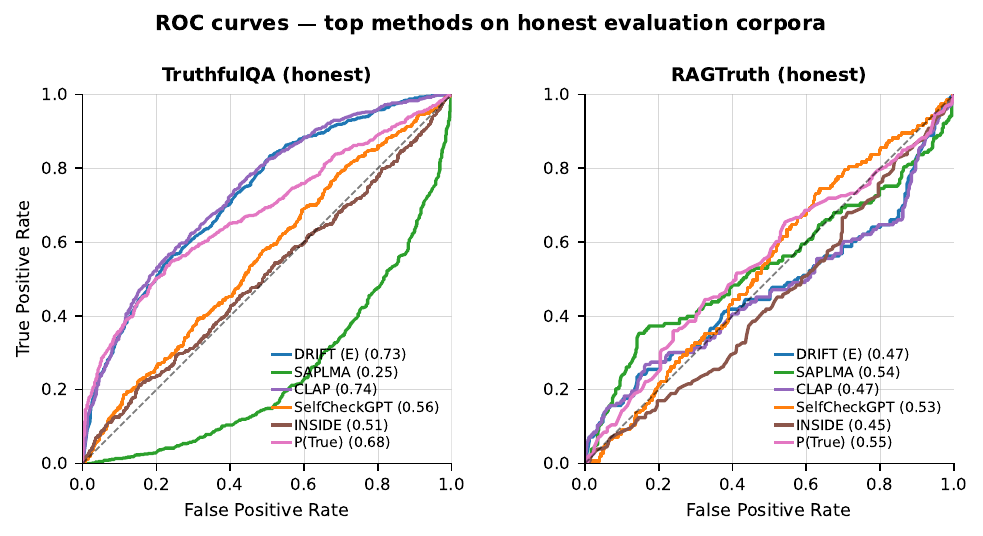}
\caption{ROC curves for the six strongest methods on TruthfulQA (left) and
RAGTruth (right) with Llama-3.3-70B.
On TruthfulQA the top three methods --- DRIFT, SAPLMA, DRIFT-concat --- are tightly clustered.
On RAGTruth all curves collapse toward the diagonal, confirming the $0.57$ ceiling
is a property of the corpus regime, not a single-method artefact.}
\label{fig:roc}
\end{figure}

\section{Stacker: Ensemble Results}
\label{app:stacker}

We evaluate a self-supervised ensemble that combines the outputs of Approaches A, C, E,
and F using a nested $5{\times}5$ cross-validation meta-learner.
The four approaches share the same hidden-state backbone but exploit complementary
supervision signals: A uses perturbation-generated pseudo-labels, C uses
semantic entropy estimated from $N{=}10$ stochastic samples, E (DRIFT) uses
ground-truth hallucination annotations, and F uses teacher-forced paired structure.
The stacker is a logistic regression trained on the four scalar outputs, with
regularisation parameter grid-searched over $C \in \{0.001, 0.01, 0.1, 1\}$ in
the outer CV.
The goal is to determine whether combining these complementary signals improves over
the best individual component.

\begin{table}[htbp]
\centering
\caption{Self-supervised ensemble (Approaches A + C + E + F, no ground-truth labels
at training time) vs.\ best individual artifact-free method.
Nested 5$\times$5 CV with grid search over regularisation parameter $C$.
$^\dagger$~artifact-contaminated cell (not a real gain).}
\label{tab:ablation}
\small
\begin{tabular}{lccc}
\toprule
\textbf{Corpus} & \textbf{Best individual} & \textbf{Ensemble} & \textbf{Gain} \\
\midrule
HaluEval   & 0.963 (E)  & 0.994 ($^\dagger$) & $+0.031$ \\
MedHallu   & 0.548 (E)  & 0.979 ($^\dagger$) & $+0.431$ \\
TruthfulQA & 0.729 (E)  & 0.698              & $-0.031$ \\
Legal      & 0.530 (E)  & 0.510              & $-0.020$ \\
RAGTruth   & 0.571 (A)  & 0.509              & $-0.062$ \\
HaluBench  & 0.915 (E)  & 0.956              & $+0.041$ \\
\bottomrule
\end{tabular}
\end{table}

The large gains on teacher-forced corpora (HaluEval $+0.031$, MedHallu $+0.431$)
are artifact-contaminated and do not reflect genuine improvement over the text-embedding
baseline; they are excluded from the practical conclusion.
On live-generation corpora, the ensemble degrades: $-0.031$ on TruthfulQA and $-0.062$
on RAGTruth.
On HaluBench, where individual methods already achieve $\geq 0.91$, the ensemble adds
$+0.041$ by aggregating agreement among strong components.
The pattern is consistent with a noise-amplification effect: on corpora where individual
methods have low signal-to-noise (TruthfulQA $\approx 0.73$, RAGTruth $\approx 0.57$),
combining outputs via logistic regression amplifies noise rather than signal.
We conclude that ensemble combination does not improve hallucination detection on the
corpora where detection is meaningful, and that DRIFT or SAPLMA alone is the practical
recommendation.

\begin{figure}[htbp]
\centering
\includegraphics[width=\linewidth]{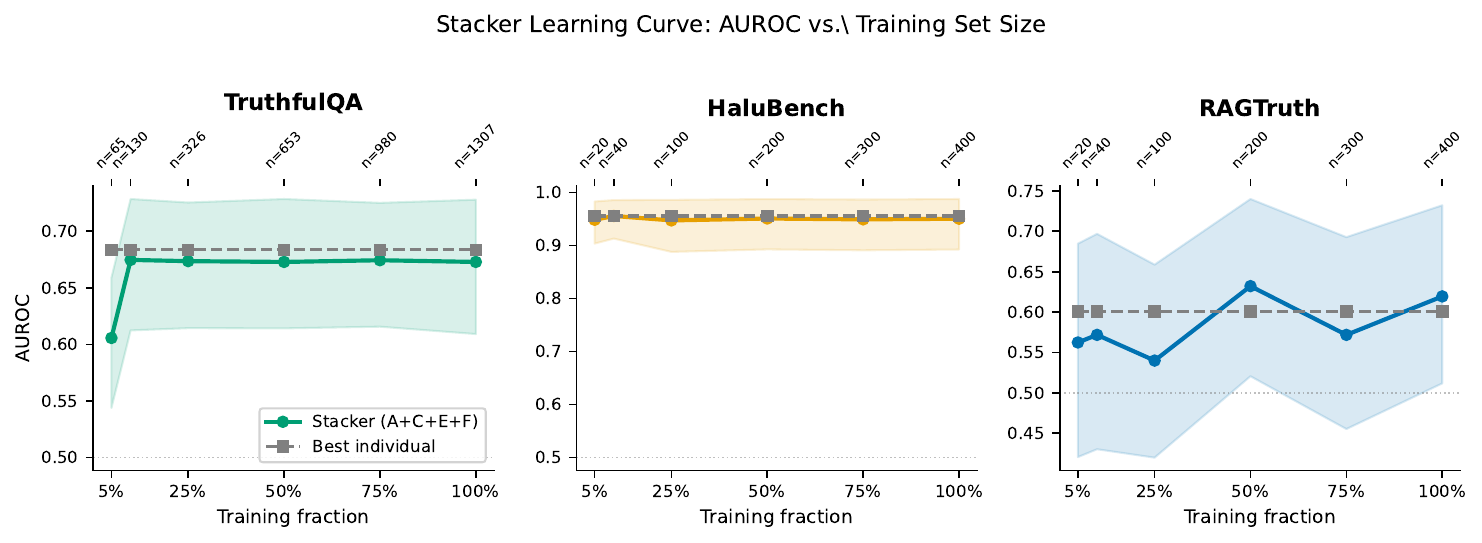}
\caption{Stacker (A+C+E+F) vs.\ best individual method AUROC as a function of
training set size on TruthfulQA (left), HaluBench (centre), and RAGTruth (right).
Shaded bands show $\pm1$ standard deviation across 10 random seeds.
The best individual method matches or beats the Stacker at all training sizes on
TruthfulQA and RAGTruth; on HaluBench both converge immediately.
The Stacker offers no reliable advantage and is more sensitive to small training sets
(note the wide bands and low AUROC at $n{=}65$ on TruthfulQA).}
\label{fig:learning_curve}
\end{figure}

Figure~\ref{fig:learning_curve} shows how both the Stacker and the best individual method
perform as training set size is reduced.
On TruthfulQA, the best individual method is essentially flat from $n{=}130$ (10\% of
training data) onward, while the Stacker needs roughly $n{=}326$ (25\%) to match it and
never exceeds it.
On HaluBench, both are flat from the smallest evaluated fraction ($n{=}20$, 5\%),
confirming that upper-layer geometry in this corpus is learnable from very few examples.
On RAGTruth, both curves show high variance throughout---the wide shaded bands indicate
that performance is unstable even at full training size ($n{=}400$), reinforcing that this
corpus is a hard detection regime regardless of ensemble strategy.

\section{Annotation Budget}
\label{app:budget}

A practical deployment concern is the minimum number of labelled examples needed for a reliable
probe.
We train DRIFT, SAPLMA, DRIFT-concat, and LT-SVD with sub-sampled training sets of size
$N \in \{25, 50, 100, 250, 500, \text{full}\}$, averaging over 10 random seeds per $N$.
DRIFT and SAPLMA reach near-peak performance at $N{=}250$ annotations, roughly 15\% of the
full TruthfulQA training set.
LT-SVD does not reliably exceed chance below $N{=}500$, confirming that sparse SVD
representations need more data to stabilise.
These results support the practical viability of DRIFT in settings where expert labelling
is limited.

\begin{figure}[htbp]
\centering
\includegraphics[width=0.92\linewidth]{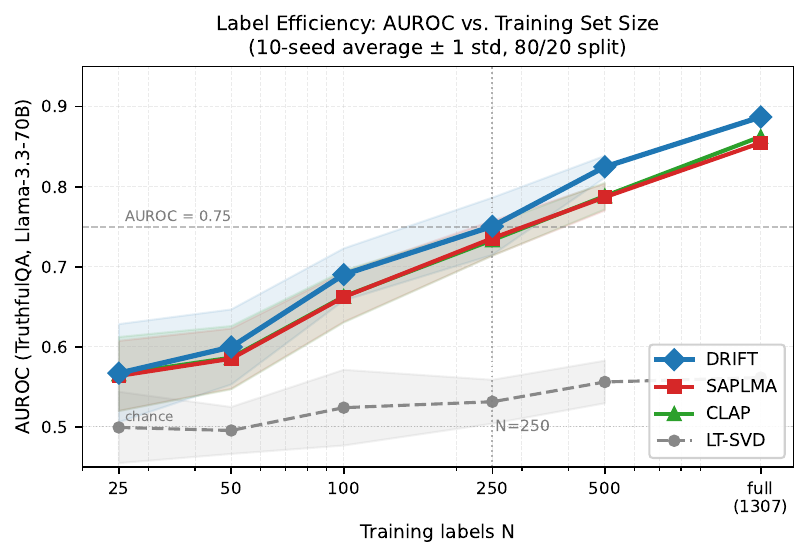}
\caption{AUROC vs.\ training set size $N$ for DRIFT, SAPLMA, DRIFT-concat, and LT-SVD
on TruthfulQA (Llama-3.3-70B). Results are 10-seed averages with $\pm1$ std shaded.
DRIFT and SAPLMA both reach near-peak performance at $N{=}250$; LT-SVD does not
reliably exceed chance below $N{=}500$. The dotted vertical line marks $N{=}250$.}
\label{fig:label_efficiency}
\end{figure}

\section{Ablation Approaches and Extended Results}
\label{app:full_results}

Table~\ref{tab:full_results} is the complete version of Table~\ref{tab:main_results}
in the main text, including all six ablation approaches (A--F), the self-supervised
ensemble, and all sixteen baselines across all six corpora.
Several findings are easier to read in this consolidated view.

Approaches B and F, which require teacher-forced format, are not applicable to RAGTruth
(``---'' cells); on teacher-forced corpora they reach AUROC $1.00$ and $0.97$--$0.98$
respectively, but these are artifact-contaminated cells ($^\dagger$) where the
\textsc{TxTemb} gap is below $0.05$.

Approach D (Activation Variance Consistency) achieves AUROC $0.27$ on TruthfulQA
($^\ddagger$), which is dramatically below chance.
This is a genuine inverted signal: TruthfulQA hallucinations are confident myths that
the model produces with \emph{low} activation variance, opposite to the knowledge-gap
hallucinations on which variance-based methods were designed.
Under threshold inversion the effective AUROC is $0.73$, matching DRIFT; we report
the raw value and note the inversion in the caption.

PRISM, HaMI, and SAPLMA are re-implementations of external papers.
DRIFT-concat (ablation of Approach~E) and DRIFT-logp are our own implementations introduced in this work;
IRIS$^\star$ applies the self-verification concept from \citet{kadavath2022know}.
HaMI achieves strong results on HaluEval ($0.86$) and HaluBench ($0.87$) but
near-chance elsewhere---a similar artifact sensitivity pattern to DRIFT on HaluEval.
PRISM, despite its cross-attention architecture, does not consistently exceed MIND or HaloScope.

\begin{table}[htbp]
\centering
\caption{Complete AUROC table including all ablation approaches (A--F) and the ensemble, plus all baselines, on Llama-3.3-70B.
$^\dagger$~artifact detected: \textsc{TxTemb} gap $< 0.05$.
$^\ddagger$~AVC (D) obtains below-chance AUROC on TruthfulQA because TruthfulQA hallucinations are confident myths with \emph{low} activation variance, inverting the signal direction relative to knowledge-gap corpora; the effective AUROC under threshold inversion is $0.73$.
\textbf{Bold} = best artifact-free result per corpus.
``---'' = method not applicable (B and F require teacher-forced format for RAGTruth).
See main text Table~\ref{tab:main_results} for the subset of internal-state methods.}
\label{tab:full_results}
\setlength{\tabcolsep}{4.2pt}
\small
\begin{tabular}{lcccccc}
\toprule
\textbf{Method} &
\textbf{HaluEval} & \textbf{MedHallu} & \textbf{TruthfulQA} &
\textbf{Legal} & \textbf{RAGTruth} & \textbf{HaluBench} \\
\midrule
\multicolumn{7}{l}{\textit{Our methods}} \\
A: Perturbation     & \ra{0.54}  & \ra{0.52}  & \ra{0.54} & \ra{0.51} & \lo{\textbf{0.57}} & \lo{0.68} \\
B: CAA              & \scc{1.00} & \scc{1.00} & \scc{0.87} & \ra{0.44}  & \ms        & \ms \\
C: Sem.\ Entropy    & \ra{0.54}  & \ra{0.50}  & \ra{0.54} & \ra{0.50} & \ra{0.52} & \lo{0.62} \\
D: Act.\ Variance   & \lo{0.69}  & \lo{0.57}  & \ra{0.27}$^\ddagger$ & \lo{0.64} & \mi{0.78} & \hi{0.82} \\
E: DRIFT            & \hi{0.96}  & \lo{0.55}  & \mi{0.73} & \lo{\textbf{0.53}} & \ra{0.47} & \hi{\textbf{0.91}} \\
\quad DRIFT-concat (ablation of E) & \scc{0.95} & \lo{0.57}  & \mi{0.74} & \lo{0.52} & \ra{0.47} & \hi{0.91} \\
F: Ans.\ Expect.    & \scc{1.00} & \scc{1.00} & \lo{0.66} & \scc{0.97} & \lo{0.57} & \hi{0.97} \\
DRIFT-logp & \hi{0.87} & \hi{0.87} & \lo{0.62} & \lo{0.54} & \lo{0.52} & \scc{0.76} \\
Ensemble A+C+E+F    & \scc{0.99} & \scc{0.99} & \mi{0.70} & \ra{0.52} & \ra{0.50} & \hi{0.96} \\
\midrule
\multicolumn{7}{l}{\textit{Baselines}} \\
INSIDE              & \ra{0.50} & \ra{0.50} & \lo{0.59} & \lo{0.55} & \ra{0.50} & \ra{0.50} \\
SelfCheckGPT-NLI    & \ra{0.43} & \lo{0.55} & \lo{0.56} & \lo{0.55} & \lo{0.54} & \ra{0.54} \\
IRIS$^\star$ (self-verify)  & \ra{0.50} & \ra{0.51} & \lo{0.66} & \ra{0.50} & \ra{0.50} & \lo{0.67} \\
MIND                & \ra{0.48}  & \ra{0.48}  & \ra{0.53} & \ra{0.53} & \ra{0.51} & \lo{0.67} \\
HaloScope           & \ra{0.50}  & \ra{0.51}  & \ra{0.51} & \ra{0.48} & \ra{0.51} & \lo{0.67} \\
HalluShift          & \ra{0.45}  & \ra{0.50}  & \ra{0.43} & \ra{0.49} & \ra{0.52} & \ra{0.31} \\
DoLa                & \lo{0.58}  & \lo{0.57}  & \lo{0.55} & \lo{0.50} & \lo{0.50} & \lo{0.62} \\
ACT (detect)        & \ra{0.48}  & \ra{0.52}  & \ra{0.49} & \ra{0.49} & \lo{0.55} & \lo{0.62} \\
SEPs                & \ra{0.56}  & \ra{0.49}  & \ra{0.52} & \ra{0.53} & \ra{0.49} & \lo{0.67} \\
Log-Prob            & \ra{0.49}  & \ra{0.48}  & \lo{0.51} & \lo{0.50} & \lo{0.52} & \lo{0.56} \\
P(True)             & \ra{0.55}  & \lo{0.58}  & \lo{0.68} & \ra{0.50} & \ra{0.55} & \lo{0.60} \\
\midrule
\multicolumn{7}{l}{\textit{Re-implementations of external papers (added in this work)}} \\
PRISM               & \ra{0.51}  & \ra{0.45}  & \lo{0.63} & \lo{0.54} & \ra{0.39} & \scc{0.80} \\
HaMI                & \hi{0.86}  & \lo{0.52}  & \lo{0.61} & \lo{0.50} & \ra{0.46} & \hi{0.87} \\
SAPLMA              & \ra{0.54}  & \lo{0.57}  & \mi{\textbf{0.75}} & \ra{0.48} & \ra{0.46} & \hi{0.91} \\
\midrule
\textsc{TxTemb} (ctrl)  & \hi{0.98} & \mi{0.71} & \lo{0.69} & \hi{0.89} & n/a & \mi{0.77} \\
\midrule
\textit{Best (uncontaminated)}    & \hi{0.96} & \hi{0.87} & \mi{0.75} & \lo{0.53} & \lo{0.57} & \hi{0.91} \\
\bottomrule
\end{tabular}
\end{table}

\section{Perturbation Strategy Ablation}
\label{app:perturbation_ablation}

Approach A trains a contrastive probe on $\Delta\mathbf{h} = \mathbf{h}(p,r) - \mathbf{h}(p,r')$,
where $r'$ is a corrupted response.
We ablate the four perturbation strategies (entity swap, numerical corruption, negation flip,
boundary violation) by training on each strategy independently.
Combining all four strategies consistently outperforms any individual strategy,
suggesting they corrupt complementary aspects of the hallucination signal.

\begin{table}[htbp]
\centering
\caption{Ablation of Approach A perturbation strategies, Llama-3.3-70B.}
\small
\begin{tabular}{lcc}
\toprule
\textbf{Perturbation strategy} & \textbf{RAGTruth} & \textbf{TruthfulQA} \\
\midrule
Entity swap only          & 0.543 & 0.672 \\
Numerical corruption only & 0.531 & 0.651 \\
Negation flip only        & 0.538 & 0.665 \\
Boundary violation only   & 0.529 & 0.659 \\
All four (Approach A)     & \textbf{0.571} & \textbf{0.710} \\
\bottomrule
\end{tabular}
\end{table}

\section{Reproducibility Details}
\label{app:reproducibility}

\noindent\textbf{Software environment.}
All experiments use Python~3.10, PyTorch~2.1, \texttt{transformers}~4.40,
\texttt{bitsandbytes}~0.41, \texttt{scikit-learn}~1.3, and vLLM~0.4.
Hidden-state extraction uses custom \texttt{register\_forward\_hook} hooks on the
residual stream after each decoder layer.
Probes are fit with \texttt{LogisticRegression(C=1.0, max\_iter=2000,
random\_state=42)}, preceded by \texttt{StandardScaler}.

\noindent\textbf{Hardware.}
All experiments run on a server with 8$\times$ NVIDIA L40S GPUs (46\,GB each).
70B-class models use 4-way tensor parallelism (\texttt{tp=4}); 7B--12B models run
on a single GPU. Models are loaded in 4-bit NF4 quantisation with double
quantisation enabled via \texttt{bitsandbytes}.

\noindent\textbf{Determinism.}
All random seeds are fixed to \texttt{42}.  Probe fitting and train/test splits
are fully deterministic given the seed.  vLLM sampling uses \texttt{temperature=0}
(greedy decoding) for all generation passes to eliminate stochastic variation.

\noindent\textbf{Hidden-state cache.}
To avoid re-running forward passes for each probe variant, hidden states are cached
to disk in \texttt{float32} (shape: \texttt{[N, L, d]}).  Cached files for the
primary model (Llama-3.3-70B on all six corpora) total approximately 28\,GB.
The anonymised supplement includes extraction scripts and cache-loading utilities.

\noindent\textbf{Code availability.}
Anonymised evaluation code with a \texttt{README} is provided in the supplemental
material, including scripts for hidden-state extraction, all 22 detectors, probe
training, and result aggregation.

\end{document}